%% file: acl_main.tex
\documentclass{article}
\usepackage{acl/acl}

\usepackage{times}

\usepackage{fullpage}
\usepackage{verbatim}  
\usepackage{multirow} 
\usepackage{multicol}

\RequirePackage{algorithm}
\RequirePackage{algorithmic}

\usepackage[dvipsnames]{colortbl}

\input{_includes}

\title{\rlft: Two-Player Online RL Fine-Tuning for LMs}

\author{
    Runlong Zhou\thanks{University of Washington. Part of this work done when Runlong was an intern at Microsoft Research, Redmond. Email: \texttt{vectorzh@cs.washington.edu}}
    \and
    Simon S. Du\thanks{University of Washington. Email: \texttt{ssdu@cs.washington.edu}}
    \and
    Beibin Li\thanks{Microsoft Research, Redmond. Email: \texttt{beibin.li@microsoft.com}}
}

\usepackage{minitoc}
\begin{document}

\maketitle

\doparttoc 
\faketableofcontents 

\begin{abstract}
\input{0_abs}
\end{abstract}

\input{1_introduction.tex}
\input{2_related_works.tex}
\input{3_preliminaries.tex}
\input{4_method.tex}
\input{5_exp_result.tex}
\input{6_conclusion.tex}

\section*{Acknowledgement}

We thank Adith Swaminathan, Julia Kiseleva, Ishai Menache, Chi Wang, Yin Tat Lee, and Yi Zhang from
Microsoft Research for useful discussions and support.

We also want to thank Jieyu Zhang from University of Washington for their valuable feedback.

\bibliography{ref.bib}




\newpage

\appendix

\part{Appendix}

\parttoc

\input{9_appendix}

\end{document}

%% file: _includes.tex
\usepackage{amsthm, amsmath, amssymb, bbm, mathabx}
\allowdisplaybreaks 
\usepackage{graphicx}
\usepackage{enumerate}
\usepackage{pifont}
\usepackage{booktabs}
\usepackage{multirow}
\usepackage{multicol}
\usepackage{stfloats}
\usepackage{xspace}
\usepackage{thmtools}
\usepackage{thm-restate}
\usepackage{hyperref}
\usepackage{cleveref}
\usepackage{anyfontsize} 
\usepackage{makecell}
\usepackage{hhline}
\usepackage{colortbl}
\usepackage{xpatch}
\usepackage{scalerel,stackengine}
\usepackage{sidecap}
\usepackage{mdframed}
\usepackage{setspace}

\usepackage{adjustbox}
\usepackage{graphicx}
\usepackage{subcaption}
\usepackage{float}

\makeatletter
\def\@fnsymbol#1{\ensuremath{\ifcase#1\or *\or \dagger\or \ddagger\or
  \mathsection\or \mathparagraph\or \|\or \diamond \or **\or \dagger\dagger
  \or \ddagger\ddagger \else\@ctrerr\fi}}
\makeatother

\makeatletter
\newcommand{\printfnsymbol}[1]{%
  \textsuperscript{\@fnsymbol{#1}}%
}
\makeatother

\crefname{condition}{Condition}{Conditions}

\crefname{assumption}{Assumption}{Assumptions}

\theoremstyle{definition}

\crefname{ALC@unique}{line}{lines}
\Crefname{ALC@unique}{Line}{Lines}
\newcounter{myalg}
\AtBeginEnvironment{algorithmic}{\refstepcounter{myalg}}
\makeatletter
\@addtoreset{ALC@unique}{myalg}
\makeatother

\newcommand{\abs}[1]{\left| #1 \right|}

\newcommand{\wh}[1]{\widehat{#1}}

\newcommand{\cA}{\mathcal{A}}

\newcommand{\cD}{\mathcal{D}}

\newcommand{\cL}{\mathcal{L}}
\newcommand{\cM}{\mathcal{M}}

\newcommand{\cS}{\mathcal{S}}
\newcommand{\cT}{\mathcal{T}}

\newcommand{\bbE}{\mathbb{E}}

\newcommand{\clip}{\mathsf{clip}}

\newcommand{\rlft}{\texttt{Reflect-RL}\xspace}

\newcommand{\autoexplore}{\texttt{AutoExplore}\xspace}
\newcommand{\aesandbox}{\texttt{AutoExploreSandbox}\xspace}
\newcommand{\aecopilot}{\texttt{AutoExploreCopilot}\xspace}

\newcommand{\nlenv}{\texttt{NatLangEnv}\xspace}
\newcommand{\taxi}{\texttt{DangerousTaxi}\xspace}

\newcommand{\reflectprompt}{p_{\textup{reflect}}}
\newcommand{\negprompt}{p_{\textup{negative}}}

\newcommand{\alfw}{\texttt{ALFWorld}\xspace}

\newcommand{\rebut}[1]{{#1}}

\usepackage[colorinlistoftodos, textwidth=18mm]{todonotes}

\def\shownotes{1}
\ifnum\shownotes=0
\newcommand{\todorz}[1]{}
\newcommand{\todorzout}[1]{}
\newcommand{\todossdout}[1]{}
\newcommand{\todossd}[1]{}
\newcommand{\todobblout}[1]{}
\newcommand{\todobbl}[1]{}
\else
\newcommand{\todorz}[1]{\todo[color=blue!10, inline]{\small RZ: #1}}
\newcommand{\todorzout}[1]{\todo[color=blue!10]{\scriptsize RZ: #1}}
\newcommand{\todossdout}[1]{\todo[color=red!10]{\scriptsize SSD: #1}}
\newcommand{\todossd}[1]{\todo[color=red!10, inline]{\small SSD: #1}}
\newcommand{\todobbl}[1]{\todo[color=orange!10, inline]{\small Beibin: #1}}
\newcommand{\todobblout}[1]{\todo[color=orange!10]{\scriptsize Beibin: #1}}
\fi

\usepackage{tikz}
\usepackage{pgfplots, pgfplotstable}
\pgfplotsset{compat=newest}
\usepgfplotslibrary{external}
\usetikzlibrary{arrows,positioning}
\usetikzlibrary{ calc, matrix, patterns, patterns.meta }
\usetikzlibrary{shapes,arrows,positioning, fit, tikzmark}
\usetikzlibrary{arrows}
\usepgfplotslibrary{statistics}
\usetikzlibrary{decorations.pathreplacing}
\usepgfplotslibrary{fillbetween}
\definecolor{msdarkblue}{RGB}{36,58,94}
\definecolor{msblue}{RGB}{0,120,215}
\definecolor{msgreen}{RGB}{16,124,16}
\definecolor{msred}{RGB}{216,59,1}
\definecolor{msgray}{HTML}{DFDFDF}

%% file: 0_abs.tex
As language models (LMs) demonstrate their capabilities in various fields, their application to tasks requiring multi-round interactions has become increasingly popular.
These tasks usually have complex dynamics, so supervised fine-tuning (SFT) on a limited offline dataset does not yield good performance.
However, only a few works attempted to directly train the LMs within interactive decision-making environments.
We aim to create an effective approach to fine-tune LMs with online reinforcement learning (RL) in these environments.
We propose \rlft, a two-player system to fine-tune an LM using SFT and online RL, where a frozen reflection model \rebut{(player)} assists the policy model \rebut{(player)}.
To generate data for the warm-up SFT stage, we use negative example generation to enhance the error-correction ability of the reflection model.
Furthermore, we designed single-prompt action enumeration and applied curriculum learning to allow the policy model to learn more efficiently.
Empirically, we verify that \rlft outperforms SFT and online RL without reflection.
Testing results indicate GPT-2 XL 1.56B fine-tuned with \rlft outperforms larger open-source LMs, such as Mistral 7B.
The benchmarks, dataset, and code involved in this work are publicly available.\footnote{\url{https://github.com/zhourunlong/Reflect-RL}}

%% file: 1_introduction.tex
\section{Introduction} \label{sec:intro}

\input{fig_tex/system}

Large language models (LLMs) have shown promising results in problem-solving, coding, and document retrieval \citep{mialon2023gaia}.
While performing these tasks, LLMs exhibit considerable reasoning, planning, and reflection skills, enabled by prompting techniques like ReAct \cite{yao2022react}, Reflexion \citep{shinn2023reflexion}, Chain of Thought (CoT, \citet{wei2023chainofthought}), Tree of Thoughts (ToT, \citet{yao2023tree}), and reasoning via planning \citep{hao2023reasoning}. Some recent studies \citep{magister2023teaching,mukherjee2023orca,mitra2023orca} also try to improve reasoning capabilities of smaller models to match those of advanced LLMs.

The reasoning and reflection skills enable LLMs to act as agents and interact with real-world environments \cite{durante2024interactive,cheng2023llf}, including code interpreters, embodied robotics \citep{shridhar2021alfworld,ahn2022i,tan2024true}, games \citep{park2023generative}, \rebut{and many other spaces \cite{vezhnevets2023generative}}.
This interaction ability is closely tied to reinforcement learning (RL), where agents can learn optimal behaviors through trial and error within an environment.

\subsection{Motivations} \label{sec:motivation}

This research is motivated by three distinct application domains within the same system,
which include: document querying \cite{izacard2022atlas}, database searching \citep{floratou2024nl2sql}, and coding \citep{chen2021evaluating}. In these applications, a chatbot needs to navigate in a file system to read documents, modify files, and execute code to answer users' questions. 
Central to these tasks is the chatbot's ability to \emph{autonomously explore} within a repository using system commands, such as, \texttt{ls}, \texttt{cd src/}, \texttt{cat main.py}, similar to the paradigm in \citet{yang2023intercode}.


Interactive chatbot for file systems \citep{nvidia_chatwithrtx_2024}, multi-agent frameworks \citep{wu2023autogen}, tool selection \citep{karpas2022mrkl,patil2023gorilla}, and many other industrial applications require interactive decision-making capabilities. Even if LLMs can perform these tasks, they are usually trained heavily with offline supervised learning rather than with online training within complex environments. Moreover, some recent studies have found that LLMs might not be able to correct themselves without external feedback during interactions \citep{huang2023large}. On the other hand, online RL training could enable LMs to dynamically adapt and make informed decisions beyond static datasets.

Some recent studies have incorporated RL to align LMs with human preference and to prompt LM for problem-solving (see \Cref{tab:rl_lm} for details).
\citet{szot2023large} and \citet{tan2024true} have started contemporary explorations to integrate LMs within interactive RL environments, but these pioneering studies have not fully utilized the LMs’ reasoning capabilities.
Motivated by the strength of RL and expansiveness of LLMs, our work aims to fine-tune smaller, faster, and more secure locally-operated LMs that are capable of decision-making and adaptation through \emph{reflection}, which are essential for domain-specific interactive tasks.




\subsection{Contributions}
\label{sec:contribution}

In this work, we introduce \rlft, {\bf a novel approach to dynamically improve LMs with online RL} (\Cref{fig:rlft}), applied with Markov decision processes (MDPs) for \emph{multi-step} decision making.
Most of the previous RL-LM works can be categorized into three classes (\Cref{tab:rl_lm}):
\ding{172} treating token-generation as RL, rather than considering embodied tasks, games, or interactive decision making within environments;
\ding{173} using LMs as agents to augment policy generation with additional textual information, without directly learning from the environment (gradient-free);
\ding{174} engaging primarily with single-step bandits rather than multi-step MDPs.
Our method seeks to improve multi-step decision making in textual environments by integrating techniques from RL and LMs, enabling LMs to adapt more efficiently to complex environments.
We summarize our \emph{key techniques} below.


\paragraph{Key Techniques:} \mbox{}

\noindent $\bullet$ {\bf Reflection (\Cref{sec:reflection}).} 
We distill reflection abilities for our domain-specific environment from GPT-4 \citep{openai2023gpt4} through supervised learning.
The distilled small LM is frozen and deployed as a reflection model \rebut{(player)} to assist the trainable policy model \rebut{(player)} in decision-making.
Reflection accelerates training convergence and improves test performance.

\noindent $\bullet$ {\bf Negative example generation (\Cref{sec:reflect_gen}).}
The reflection data gathered from GPT-4 is unbalanced, with the majority consisting of positive (near-optimal) decisions.
To balance the dataset, we generate negative examples by perturbing the GPT-4 trajectories and optimal trajectories.
Negative examples enhance the quality of reflection, ultimately leading to better success rates in the benchmarks.

\noindent $\bullet$ {\bf Single-prompt action enumeration (\Cref{sec:action_enum}).}
We incorporate all possible valid actions into a single prompt, allowing the LM to select the appropriate option using only one token.
This approach improves upon the normalization techniques in previous works to generate valid actions and also reduces time complexity.

\noindent $\bullet$ {\bf Task-specific curriculum learning (\Cref{sec:curriculum}).}
The core challenges of RL include planning for a long horizon and sparse reward signals.
Vanilla policy optimization methods often fail to obtain sufficient useful trajectories efficiently.
We incorporate the idea of curriculum learning into our pipeline, designing a specific curriculum to guide training by giving extra rewards or scheduling the data order.

\paragraph{New Benchmark for Online RL Fine-Tuning.}
Additionally, we introduce \autoexplore, \emph{a benchmark inspired by industrial applications}, along with other benchmarks adapted from previous works.
These benchmarks are suitable for both \emph{research} and \emph{application} purposes.
They can be integrated with either local LMs for training or remote LLMs for in-context inference.
Our demonstrations show positive results of LLMs on industrial applications.
Both RL training and data generation are made easy by their use.

\paragraph{Paper Overview.}
This paper begins by discussing LLMs in \Cref{sec:rel} and RL preliminaries in \Cref{sec:preliminaries}. 
Then, we introduce our proposed \rlft in \Cref{sec:method} and benchmarks in \Cref{sec:benchmark}.
The results are presented in \Cref{sec:exp_results}.
Finally, we discuss the findings and future directions in \Cref{sec:disc_conc}.

%% file: fig_tex/system.tex
\begin{figure*}
    \centering
\begin{tikzpicture}[node distance=2cm, auto,
block/.style={
  rectangle,
  draw,
  text width=2.5cm,
  text centered,
  rounded corners,
  minimum height=2.5cm,
},
line/.style={
  draw, 
  -latex',
  shorten >=2pt,
  line width=0.3mm
},
train_line/.style={
  draw, red,
  -latex',
  shorten >=2pt,
  line width=0.4mm,
  dashed,
  dash pattern=on 6pt off 3pt on 1pt off 3pt,
  text=black
},
cloud/.style={
  draw,
  ellipse,
  minimum height=2em
}]

\tikzstyle{fitbox} = [rectangle, dashed, draw=black, inner sep=0.5cm]

\node [block, draw=none] (env) {\includegraphics[width=1.8cm]{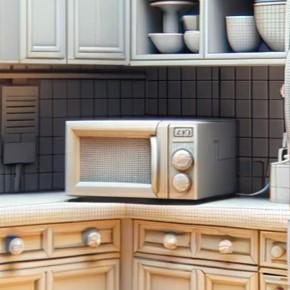} \\ Environment};

\node [block, right=0.5cm of env, yshift=1.5cm] (obs) {Observation\\
\vspace{0.2cm}
  \parbox{1.8cm}{ 
  \centering
  \tiny  You are in the middle of a room. You see a cabinet 18, a cabinet 3, a ...}};

\node [block, right=0.5cm of env, yshift=-1.5cm] (actions) {Possible Actions
\\~\\
\vspace{0.2cm}
  \parbox{1.8cm}{ 
  \centering
  \tiny  1. Close microwave\\2. Put tomato in microwave\\3. Go to cabinet\\ ...}
  
};

\node [block, circle, minimum height=1cm, text width=1.5cm, right=0.55cm of obs, yshift=0cm, draw=none, fill=cyan!20] (reflect-agent) {\includegraphics[width=0.8cm]{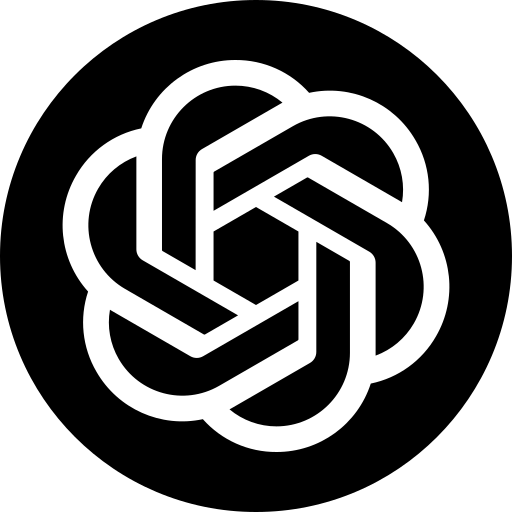}\includegraphics[width=0.8cm]{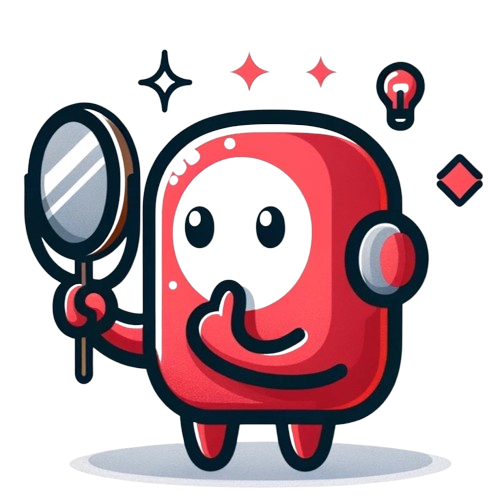}\\ Reflect Agent};

\node [block, right=0.5cm of reflect-agent, fill=yellow!30] (reflection) {Reflection \\
  \centering
  \vspace{0.2cm}
  \parbox{2.3cm}{ 
    \tiny 
    \setlength{\baselineskip}{8pt} 
    Taking the tomato to the microwave is correct. Next step is to place the tomato ...
  }
};
\node [block, right=0.49cm of actions, fill=blue!20] (memory) {\includegraphics[width=0.5cm]{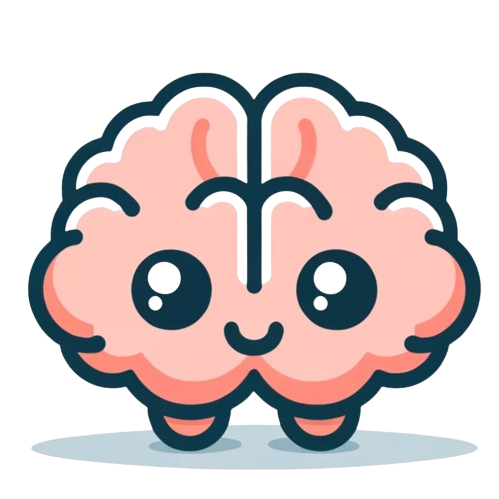} \textbf{Memory}\\
  \vspace{0.2cm}
  \parbox{2.5cm}{ 
  \centering
  \tiny \textbf{\tiny \textcolor{red}{Task: Cook Tomato}}\\
  Act 0: Go to fridge\\
  Obs 0: It is closed \\
  ...\\
  Act 4: Open microwave\\
  Obs 4: You see a mug ...
  }
};

\node [block, right=0.2cm of reflection, draw=none] (database) {\includegraphics[width=1.9cm]{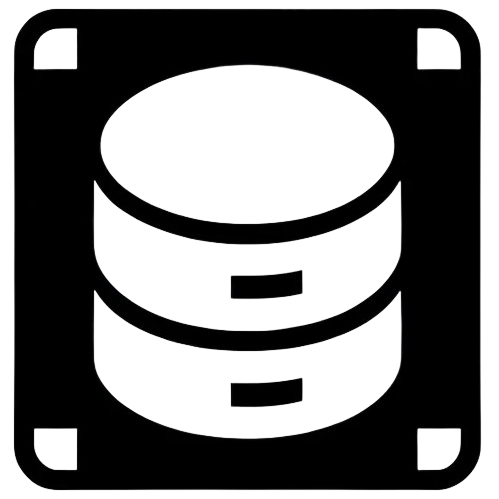}\\Database};

\node [block, circle, minimum height=1cm, text width=1.5cm, right=0.5cm of memory, draw=none, fill=orange!20] (policy-agent) {\includegraphics[width=0.8cm]{fig_tex/src/chat-gpt.png}\includegraphics[width=0.8cm]{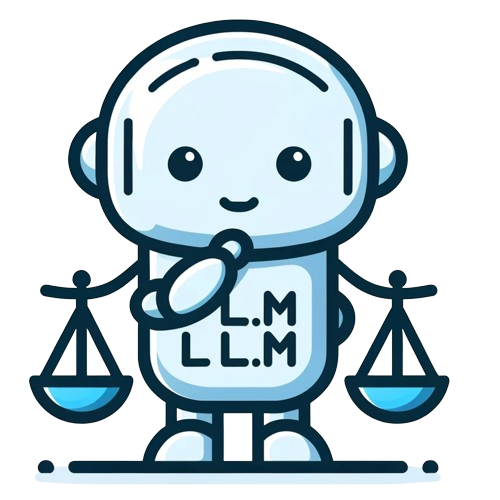} \\ Policy Agent};

\node [right=0.65cm of policy-agent, fill=green!20, rounded corners] (policy-action) {\includegraphics[width=0.3cm]{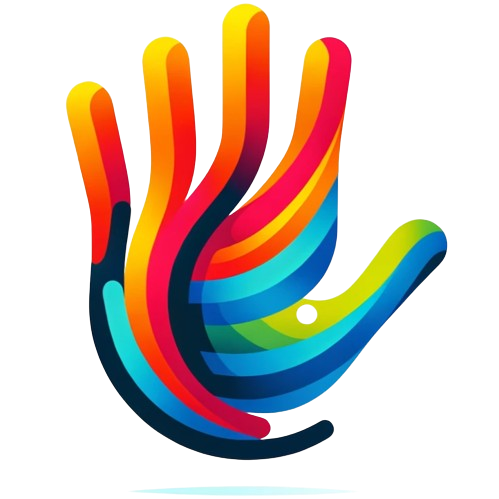} Action: 2};


\path [line] (env) -- (obs);
\path [line] (env) -- (actions);
\path [line, <->] (memory) -- (reflect-agent);
\path [line] (obs) -- (reflect-agent);
\path [line] (reflect-agent) -- (reflection);
\path [line] (reflection) -- (policy-agent);
\path [line] (actions.south) -- ++(0, -0.5) -| (policy-agent.south);
\path [line,shorten >=-4mm] (reflection) -- (database);

\path [line] (policy-agent.east) -- (policy-action);
\path [line] (policy-action.east) -- ++(0.3, 0) -- ++ (0, 4.5) -| (env.north);
\path [line] (policy-action) --  (database.south);

\path [train_line,shorten <=-4mm] (database.east) -- ++(0.2, 0) -- ++ (0, 1.8) -|  (reflect-agent.north) node[midway, above, dashed, rounded corners, fill=red!20,  xshift=0.8cm, yshift=0.1cm] {1. SFT};

\path [train_line,shorten <=-4mm] (database.east) -- ++(0.2, 0) |- ([yshift=-0.4cm]policy-agent.east) node[midway, above, dashed, rounded corners, fill=red!20,  xshift=-2.3cm, yshift=-0.7cm] {1. SFT};

\path [train_line] (env.south) -- ++(0, -2.2) -- ++ (1.0, 0.0) -| ([xshift=0.3cm]policy-agent.south) 
node[midway, above, rounded corners, xshift=-8.5cm, yshift=0.1cm] {[ Reward: +1 ]}
node[midway, above, fill=red!20, rounded corners, xshift=1.6cm, yshift=0.1cm] {2. Online RLFT};

\end{tikzpicture}

   \caption{\rlft Pipeline. Solid lines represent the forward pass for both data generation and inference. Agents (in circular nodes) are language models capable of generating reflections and making decisions. Red dashed lines represent the loss and gradient calculation during the training periods: the reflection agent is trained with SFT, while the policy agent is trained first with SFT and then with online RLFT.
   \rebut{Detailed illustrations for each stage can be found in \Cref{sec:illustration_pipeline}.}
   }
    \label{fig:rlft}
\end{figure*}

%% file: 2_related_works.tex
\section{Related Works} \label{sec:rel}


\paragraph{Language models (LMs).}
LMs play a pivotal role in tasks such as sentiment analysis \citep{zhong2023chatgpt,wang2023chatgpt}, machine translation \citep{GULCEHRE2017137,lample2019crosslingual}, and automated text generation \citep{chen2020distilling,dathathri2020plug}, showcasing their versatility and capability in handling complex linguistic structures.

\paragraph{LM agents and multi-agent collaborations.}
Autonomous LM agents \citep{bran2023chemcrow,park2023generative,wu2023autogen,wang2023survey} underscore LMs' capabilities of autonomous and collaborative problem-solving.
Such agent collaboration can achieve a level of sophistication and efficiency that is difficult to obtain through solo efforts.

\paragraph{Fine-tuning of LMs.}
Supervised fine-tuning (SFT, \citet{howard2018universal,Radford2019LanguageMA}) and reinforcement learning from human feedback (RLHF) are the most commonly used alignment methods for adapting pre-trained LMs to specific tasks. 
Additionally, LoRA \citep{hu2021lora}, QLoRA \citep{dettmers2023qlora}, and other parameter-efficient fine-tuning (PEFT) algorithms can facilitate this process.

\paragraph{LMs for interactive decision-making.}
As summarized in \Cref{tab:rl_lm} and discussed in \Cref{sec:contribution}, only a few studies have applied online RL to LMs for making multi-step decisions.
\citet{szot2023large} and \citet{tan2024true} are the two most relevant studies.



\begin{table*}[ht]
    \centering
    \begin{tabular}{c|c|c|c|c|c}
        \hline
        {\bf Category} & {\bf Works} & \scriptsize \makecell{{\bf Direct}\\{\bf Interaction}} & \scriptsize \makecell{{\bf Bandit}\\{\bf or MDP}} & \scriptsize {\bf Reflection} & \scriptsize \makecell{{\bf Training}\\{\bf Method}} \\
        \hline 
        \hline
        \makecell{Token-\\generation\\as RL} & \makecell{\citet{lu2022quark}, \citet{ramamurthy2023reinforcement},\\ \citet{luong2024reft}, \citet{yuan2024self}} & {\bf Yes} & {\bf MDP} & No & {\bf RL} \\
        \hline
        \multirow{4}{*}{\makecell{LMs as\\agents}} & \makecell{\citet{park2023generative}, \citet{zhang2023large}, \\ \citet{shinn2023reflexion}}& {\bf Yes} & \multirow{4}{*}{\bf MDP} & \multirow{3}{*}{\bf Yes} & \multirow{4}{*}{None} \\
        \cline{2-3}
        & \makecell{\citet{huang2022inner}, \citet{yao2022react}, \\ \citet{yao2023retroformer}, \citet{du2023guiding}} & \multirow{2}{*}{No} & & & \\
        \cline{2-2}\cline{5-5}
        & \makecell{\citet{ahn2022i}} & & & No & \\
        \hline
        RLHF & \makecell{\citet{ziegler2020finetuning}, \citet{stiennon2022learning},\\ \citet{bai2022training}, \citet{ouyang2022training}} & {\bf Yes} & Bandit & No & {\bf RL} \\        %
        \hline
        SFT & \citet{shridhar2021alfworld} & {\bf Yes} & {\bf MDP} & No & Supervised \\
        \hline
        \multirow{2}{*}{\makecell{RL \\ Fine-tuning}} & \makecell{\citet{szot2023large}, \citet{tan2024true}} & \multirow{2}{*}{\bf Yes} & \multirow{2}{*}{\bf MDP} & No & \multirow{2}{*}{\bf RL} \\
        \cline{2-2} \cline{5-5}
        & {\bf This work} & & & {\bf Yes} & \\

        \hline 
        
    \end{tabular}
    \caption{Comparison between works involving LMs and RL.
    ``Direct interaction'' indicates whether the LM plays the role as the policy model directly interacting with the environment, so a ``No'' means it plays indirectly by assisting another non-language policy model.
    ``Bandit or MDP'' indicates whether the environment is a single-step bandit or a multi-step MDP.
    ``Reflection'' indicates whether this work elicits the reasoning ability of the language model to generate reflections and help with planning in RL.
    ``Training method'' indicates whether the LM is being trained and if yes, the method.}
    \label{tab:rl_lm}
\end{table*}

%% file: 3_preliminaries.tex
\section{Preliminaries} \label{sec:preliminaries}

\paragraph{Notations.}
For any set $\mathcal{X}$, we use $\Delta(\mathcal{X})$ to denote the probability simplex over $\mathcal{X}$.
Let the tokenizer be fixed throughout the paper.
For a string $s$, we use $\abs{s}$ to denote the number of tokens in $s$ after using this fixed tokenizer.

\paragraph{Markov decision processes (MDPs).}
Reinforcement learning (RL, \citet{rl_sutton_barto}) problems are usually formulated as MDPs.
They enable agents to learn optimal behaviors through interacting with the environment, without human intervention or labeling.
A (finite-horizon) MDP can be described as $\cM=(H,\ \cS,\ \cA,\ \mu,\ \cT,\ r)$, where $H$ is the planning horizon, $\cS$ is the state space, and $\cA$ is the action space.
$\mu\in \Delta(\cS)$ is the initial state distribution, which can represent a distribution over tasks.
We study \emph{deterministic environments} in this work as the tasks in our motivations are deterministic.
The transition function maps a state-action pair to a state $\cT: \cS\times \cA\rightarrow \cS$, and the reward function immediately yields a reward $r: \cS\times \cA\rightarrow [-1,1]$.
Given a (Markovian) policy $\pi: \cS \to \Delta(\cA)$, we define its value function and $Q$-function as
\begin{gather*}
    V_h^\pi (s) := \bbE_\pi \left[\left. \sum_{t = h}^H r_t\ \right|\ s_h = s \right], \\
    Q_h^\pi (s, a) := \bbE_\pi \left[\left. \sum_{t = h}^H r_t\ \right|\ (s_h, a_h) = (s, a) \right].
\end{gather*}
The expected return of a policy $\pi$ is $J^\pi := \bbE_{s_1 \sim \mu} [V_1^\pi (s_1)]$, and the goal of RL is to find the optimal policy maximizing $J^\pi$.

When modeling an application as an MDP, we may encounter the fact that each state $s$ has a separate ``valid'' action space $\cA (s)$.
Though we can define $\cA = \cup_{s \in \cS} \cA (s)$, the union could be intractably large.
A viable workaround is to define a mapping function $f_s$ at each state, \rebut{such that $\cA (s) \subseteq \{ f_s(a)\ |\ a \in \cA \}$. This formulation works smoothly with our approach named ``single-prompt action enumeration'' (\Cref{sec:action_enum}) where $\cA$ consists of choices such as 0, 1, 2, \ldots, and $f_s (a)$ maps them to detailed actions.}

\paragraph{Policy optimization for MDPs.}
Policy optimization is an approach to solve MDPs using parameterized policies.
Policy optimization techniques for MDPs surround the class of policy gradient (PG, \rebut{or REINFORCE algorithm}, \citet{NIPS1999_464d828b}) methods, which directly adjust the parameters of the policy in a way that maximizes $J^\pi$.
Let $\pi_\theta$ be a policy parameterized by $\theta$, then the policy gradient is computed as
\begin{align*}
    \nabla_\theta J^{\pi_\theta}
    = \sum_{h = 1}^H \bbE_{s, a \sim d_h^{\pi_\theta}} \left[ Q_h^{\pi_\theta} (s, a) \nabla_\theta \ln \pi_\theta (a | s) \right].
\end{align*}
\rebut{Here $d_h^{\pi_\theta}$ is the distribution of $(s, a)$ pairs at step $h$ under policy $\pi_\theta$.}
An update step using policy gradient is $\theta_{t + 1} = \theta_t + \eta \nabla_\theta J^{\pi_{\theta_t}}$.

Proximal Policy Optimization (PPO, \citet{schulman2017proximal}) is another exemplary method applied in this field, whose details are deferred to \Cref{sec:ppo}.

%% file: 4_method.tex
\section{Methodology} \label{sec:method}

\subsection{\rlft}

Here, we propose \rlft, an online reinforcement learning fine-tuning method for LMs in MDPs.

\subsubsection{LM as an RL policy}

We use a language model as an RL policy $\pi_\theta (a | s)$ where $s = (s_1, s_2, \ldots, s_L) \in \cS$ is the current state (represented by tokens) and $a = (a_1, a_2, \ldots, a_K) \in \cA(s)$ is the generated token sequence (also represented by tokens).
Let $a_{:k}$ denote the subsequence $(a_1, a_2, \ldots, a_k)$.
We apply policy model to multi-step RL tasks, where
the language model reads $s$ in the input prompt, and then generate $a$ in the completion.


In environments where states are not represented in natural languages, we need a function $p (s)$ to convert the original state $s$ to make it a legal input for an LM. For instance, $p$ can be a ViT \citep{dosovitskiy2020image} for images, as used in LLaVA \citep{liu2023visual}); or, $p$ can be a text representation for simple graphs.
Naturally, for $s_1 \ne s_2$, we require $p(s_1) \ne p(s_2)$.
With a little bit abuse of notations, prompt $p (s)$ and state $s$ are equivalent throughout our paper.

\subsubsection{Training stages of \rlft}

We propose a two-stage training pipeline for the above-mentioned language model policy.
An illustration is shown in \Cref{fig:rlft}.

\paragraph{Stage 1. Supervised fine-tuning (SFT).}
The tasks included in this work all require the instruction-following capability to a certain degree: for any valid state $s$, the generated action $a$ should follow an instructed format.
For example, the model should output a paragraph reflecting on previous decisions before making the next action, with two parts separated by a special token.
For these tasks, we fine-tune LMs with a dataset $\cD$ comprised of strings which follow the instruction.
This process only calculates losses on the completion part.

\paragraph{Stage 2. Reinforcement learning fine-tuning (RLFT).}
We use reinforcement learning to fine-tune a pretrained language model $\pi_{\theta_0}$, which can either be a publicly available LM or the one after SFT.
This stage proceeds in $T$ update steps.
In step $t \in \{0, 1, \ldots, T - 1\}$, we use $\pi_{\theta_t}$ to sample a batch of $B$ trajectories from the environment, estimate $Q$-functions for each step, then perform updates using the policy optimization algorithm.


\subsubsection{Training details} \label{sec:reflection}

\paragraph{Reflection-aided decision-making.}
As demonstrated in previous works \citep{yao2022react,yao2023retroformer,shinn2023reflexion}, generating reflection is helpful for improving the decision-making performance, which inspires us to incorporate reflection in RL.
We combine the idea of reflection with both SFT and RLFT.
Specifically, we first assume access to an independent reflection model $R$ to generate reflections before the policy model $\pi_\theta$ makes decisions.
Upon observing state $s$, $R$ generates the reflection $R = R(s)$ which possibly includes analyses of current situation and plans of future steps.
Then, the policy model generates the action after taking both $s$ and $R$ as inputs.
The reflection model $R$ is independent of $\pi_\theta$: it can be either a local, pretrained language model, or a publicly-hosted LLM such as GPT-4 or Gemini \citep{geminiteam2023gemini}.
One illustration can be found in \Cref{sec:illustration_radm}.

In our work, we train local LMs in SFT stage using data collected from Azure OpenAI GPT-4 (details in \Cref{sec:dataset}) to serve as the reflection model $\wh{R}_\phi$ (\Cref{alg:line:spt_ref} of \Cref{alg:rlft}).
$\wh{R}_\phi$ is frozen (denoted as $\wh{R}$) throughout the RLFT stage.
The policy model is SFTed using data containing the reflection (\Cref{alg:line:spt_policy} of \Cref{alg:rlft}).
Formally, let $\cD = \{(s_i, R_i, \alpha_i, a_i)\ :\ 1 \le i \le N\}$ be the dataset (\Cref{alg:line:dataset} of \Cref{alg:rlft}), with $\abs{R_i} = L_i$ and $\abs{a_i} = K_i$, then the loss functions are
\begin{align*}
    & \cL_{\textup{reflect}} (\phi) = \frac{1}{N} \sum_{i = 1}^N \sum_{j = 1}^{L_i} -\log \wh{R}_\phi (R_{i, j} | s_i, R_{i, :j - 1}), \\
    & \cL_{\textup{policy}} (\theta) = \frac{1}{N} \sum_{i = 1}^N \sum_{j = 1}^{K_i} -\log \pi_\theta (a_{i, j} | s_i, R_i, \alpha_i, a_{i, :j - 1}).
\end{align*}
Here $\alpha_i = \alpha (\cA (s_i))$ and $\alpha$ is the action enumeration function defined in \Cref{sec:action_enum}.

In RLFT stage, we first query $\wh{R}$ for the reflection, then incorporate this reflection into the policy model's input (\Cref{line:alg:rl_ref,line:alg:rl_policy} of \Cref{alg:rlft}).
The probability of the action is
\begin{align*}
    \pi_{\theta_t} (a | s) = \prod_{j = 1}^{K} \pi_{\theta_t} (a_j | s, \wh{R}, \alpha, a_{:j - 1}).
\end{align*}

\paragraph{\rebut{Two-player} design simplifies the training process.} Splitting responsibilities to two \rebut{players} (reflection and policy) can simplify the RLFT stage because the gradients of the policy model do not affect the reflection model.
We experimented using the same model for reflection and policy, while computing gradients only on the policy part.
Observations (in \Cref{sec:same_model}) show that such implementation greatly degraded the reflection ability.
An alternative \rebut{single-player} approach is to perform RL and SFT concurrently so that \rebut{the reflection ability can be retained}, but this strategy would complicate the training process.

\begin{algorithm}[t!]
\caption{Training with \rlft \label{alg:rlft}}
\begin{algorithmic}[1] \small
    \STATE{\textbf{Input and initialize:} Environment $E$, batch size $B$, prompting function $p$, action enumeration function $\alpha$, SFT data size $N$, pretrained LM $\cM$, LLM to generate reflection data $\cM_R$, number of updates $T$.}
    
    \STATE{$\cD_{\textup{reflect}} \gets \varnothing$, $\cD_{\textup{negative}} \gets \varnothing$.}
    \FOR {$n=1,2,\ldots,N$}
        \STATE {$E$.reset(), $h \gets 1$}
        \WHILE {$\lnot E$.done}
            \STATE {$s_h \gets E$.observation()}
            \STATE {$R_h \gets \cM_R (p (s_h), \reflectprompt)$} \label{line:alg:ref_gen_start}
            \STATE {$a_h \gets \cM_R (p (s_h), R_h, \alpha(\cA (s_h)))$} 
            \STATE {$\cD_{\textup{reflect}} \gets \cD_{\textup{reflect}} \cup \{(s_h, R_h, \alpha(\cA (s_h)), a_h)\}$} \label{line:alg:ref_gen_end}
            \STATE {$a_{h}' \sim $ Uniform$(\cA (s_{h}) \backslash a_{h})$ ~~~// \textcolor{gray}{random action}}
            \STATE {$E$, $E' \gets E$.step$(a_h)$, $E$.step$(a_{h}')$}
            \STATE{$h \gets h + 1$}
            \STATE{\textcolor{gray}{// Look ahead: reflect after the ``wrong" action}}
            \STATE {$s_{h}' \gets E'$.observation()}
            \STATE {$R_{h}' \gets \cM_R (p (s_{h}'), \negprompt)$}
            \STATE {$a_{h}' \gets \cM_R (p (s_{h}), R_{h}, \alpha(\cA (s_{h}')))$}
            \STATE {$\cD_{\textup{negative}} \gets \cD_{\textup{negative}} \cup \{(s_{h}', R_{h}', \alpha(\cA (s_{h}')), a_{h}')\}$}
        \ENDWHILE
    \ENDFOR


    \STATE {$\cD \gets \cD_{\textup{reflect}} \cup \cD_{\textup{negative}}$} \label{alg:line:dataset}
    \STATE {$\wh{R} \gets $ SFT$(\cM, \{ (R\ |\ p(s)) \in \cD \})$} \label{alg:line:spt_ref}
    \STATE {$\pi_{\theta_0} \gets $ SFT$(\cM, \{ (a\ |\ p(s), R, \alpha(\cA (s))) \in \cD \})$} \label{alg:line:spt_policy}
    
    \FOR {$t=0,1,\ldots, T - 1$}
        \FOR {$b=1, 2, \ldots, B$}
            \STATE {$E$.reset(), $h \gets 1$}
            \WHILE {$\lnot E$.done}
                \STATE {$s_h \gets E$.observation()}
                \STATE {$R_h \sim \wh{R} (p (s_h))$} \label{line:alg:rl_ref}
                \STATE {$a_h \sim \pi_{\theta_t} (p (s_h), R_h, \cA (s_h))$} \label{line:alg:rl_policy}
                \STATE {$E \gets$ $E$.step$(a_h),\ h \gets h + 1$}
            \ENDWHILE
            \STATE {$\tau_b \gets (s_1, R_1, \cA (s_1), a_1; \ldots)$}
        \ENDFOR
        \STATE{$\theta_{t + 1} \gets $Policy\_Gradient$(\theta_t, \{\tau_1, \ldots, \tau_B \})$}
    \ENDFOR
\end{algorithmic}
\end{algorithm}

\subsection{Generating Reflection for Training} \label{sec:reflect_gen}
Two components are essential in reflection generation:

\noindent $\bullet$ {\bf Logical consistency.}
We want a trajectory to be logically consistent, in that the action $a_h$ at step $h$ logically follows the reflection $R_h$ at step $h$.
This requirement is critical for the policy model $\pi_\theta$ to derive the correct action from the reflection.

\noindent $\bullet$ {\bf Negative examples.}
Using optimal or oracle actions to train policy models is a well-established strategy in RL.
However, employing this strategy to generate training data with LLM may introduce a bias towards producing predominantly affirmative \emph{reflections} on previous actions.
If such data are exclusively used for training, the reflection model might merely flatter the decisions made by the policy model, without providing substantive self-reflections.
Consequently, the model’s ability to generalize to new or sub-optimal actions could be significantly limited.
To mitigate this, incorporating negative examples (sub-optimal actions) can help balance the dataset and enhance the error-correcting capabilities of the reflection model.

Accordingly, we use two methods to generate the SFT dataset, with two types of special prompts $\reflectprompt$ and $\negprompt$.

At step $h$, we get the state $s_h$ from the environment and send $(s_h, \reflectprompt)$ to GPT-4.
Here $\reflectprompt$ tells GPT-4 to first analyze current situation, plan for the next steps, then generate the action.
GPT-4 will generate a response, from which we can easily extract out reflection $R_h^{\textup{GPT}}$ and action $a_h$ because of GPT-4's high-level instruction-following capability.
Next we send $a_h$ to the environment and increment $h$ until termination.
The above procedure generates a \emph{logically consistent} trajectory $\tau$.
The illustration can be found between \Cref{line:alg:ref_gen_start,line:alg:ref_gen_end} of \Cref{alg:rlft} and \Cref{fig:rlft}.

To get negative examples, we start from $\tau$ or an optimal trajectory $\tau^\star$ by perturbing each step.
For any step $h$, we first restore the environment to state $s_{h-1}$, then we randomly pick an action $a_{h-1}'$ from the set $\cA (s_{h-1}) \backslash \{a_{h-1}\}$.
This perturbed action will lead us into another state $s_h'$.
We send $(s_h', \negprompt)$ to GPT-4, where $\negprompt$ tells GPT-4 that the last action $a_{h-1}'$ is sub-optimal, and lets it to find out the reason of sub-optimality, plan for the next steps to correct the mistake, then generate the action $a_h'$.
The reflection generated at this step is $(R_h^{\textup{GPT}})'$.
We halt at this step, using only $(s_h', (R_h^{\textup{GPT}})', a_h')$ as a \emph{negative example}.


\subsection{Single-Prompt Action Enumeration} \label{sec:action_enum}

The action spaces in the benchmarks are extremely large and \emph{state-dependent}.
Moreover, a valid action spans over several tokens, and has constraints on the token combination.
For instance, in ALFWorld, the action spaces can differ across tasks or locations, due to variations in the objects that can be interacted with.
A typical valid action is ``\texttt{go to cabinet 10}'' which contains $4$ tokens, while ``\texttt{take cabinet 10}'' is invalid.
However, this valid action may become invalid when presented in another task where ``cabinet 10'' does not exist.
As stated in various works \citep{ahn2022i,tan2024true}, it is highly possible for the language model to generate a long token sequence that does not meet the constraints.

The remedies proposed by these works share the same spirit.
SayCan \citep{ahn2022i} and Action prompt normalization \citep{tan2024true} are similar approaches enumerating all the valid actions $a \in \cA (s)$, calculating the probability $\pi_\theta (a | s)$, and normalizing over $\cA (s)$.
Calculating $\pi_\theta (a | s)$ using a Transformer model takes $\Theta ((\abs{s} + \abs{a})^2)$ time.
This approach takes $\Theta (\sum_{a \in \cA (s)} (\abs{s} + \abs{a})^2) = \Theta (\abs{\cA (s)} \abs{s}^2 + \sum_{a \in \cA (s)} \abs{a}^2)$ time, which is intractable when $\abs{\cA (s)}$ is large.
Here we assume $\abs{s} \gg \abs{a}$ as in almost all of the benchmarks.
For two benchmarks (\autoexplore and \alfw) considered in our work, we have $\abs{\cA (s)} \approx 20$, $\abs{s} \approx 500$, and $\abs{a} \approx 5$ for almost all the states.
As a result, action prompt normalization cannot be applied to our benchmarks.

We propose \emph{single-prompt action enumeration} which shares spirit with many language classification tasks \citep{zellers2018swag,bisk2019piqa,hendrycks2021measuring} to reduce time complexity while enforcing valid actions.
This method works on two sides.
On the \emph{environment} side, we introduce an extra component: the action enumeration function $\alpha$.
Suppose $a_1, a_2, \ldots$ is an order of actions in $\cA (s)$, then we compose $\alpha (\cA (s)) = (1, a_1; 2, a_2; \ldots)$ by explicitly writing down the choice letter $i$ and action $a_i$.
$\alpha$ is sent to the policy model as additional input, together with state $s$ and reflection $R$.
On the \emph{model} side, we restrict the policy model to output exactly \emph{one} token, representing the choice in $\alpha$.
We also mask out rows of \texttt{lm\_head} (neurons of the final output layer) that does not decode into a choice letter.
With these combined, we are ensured that the generated action is valid.
As a comparison with action prompt normalization, the running time of our approach is $\Theta ((\abs{s} + \sum_{a \in \cA (s)} \abs{a})^2) = \Theta (\abs{s}^2 + \sum_{a \in \cA (s)} \abs{a}^2)$, which is strictly better.
Here reflection $R$ is considered as part of $s$ without loss of generality.

\subsection{Curriculum Learning} \label{sec:curriculum}

Curriculum learning \citep{ELMAN199371,10.1145/1553374.1553380} is a paradigm in machine learning using a \emph{topological ordering} of tasks to help with training.
Starting with easy tasks, the model can have a faster convergence on hard tasks compared with directly training on them.
In this work, we experiment on a curriculum design called ``extra reward signal''.
For tasks with long horizons and sparse rewards, it is nearly impossible for a policy to sample a trajectory with a meaningful reward signal, thus policy gradient methods will make slow progress.
We design the curriculum by manually adding rewards to some ``milestones''.
In experiments of \taxi (see \Cref{sec:benchmark}), which requires to first pick up then drop off a passenger while only giving reward after a successful dropoff, we design the curriculum to give a reward after a successful pickup.




%% file: 5_exp_result.tex
\input{fig_tex/rst_table}

\section{Benchmarks}
\label{sec:benchmark}

Motivated by the LLF-Bench \cite{cheng2023llf}, we have created a natural language environment base class (\nlenv) that is compatible with the OpenAI Gym framework, characterized by its unique approach of utilizing textual representations for both observations and actions. This adjustment allows us to effectively train and test language models.


\paragraph{\autoexplore.}

To verify our methodology of \rlft on the exploration example mentioned in \Cref{sec:motivation}, we built a complete benchmark for autonomous exploration.
This benchmark contains three components: a \aesandbox for file protection, a multi-agent system \aecopilot for interactive decision-making, and a labeled dataset for performance assessment.
The \autoexplore environment enables LMs to interact with the file system safely, with the ultimate goal of answering a natural language question specified by users.
The labeled dataset is composed of several real-world and synthesized repositories, with over 2500 trajectories.
See \Cref{sec:ae_details} for more details.

This exploration task draws inspiration from Retrieval Augmented Generation (RAG) \cite{lewis2020retrieval}  and InterCode \cite{yang2023intercode}. RAG's performance is linearly dependent on the amount of content (e.g., number of files) in the search space, presenting scalability challenges. In contrast, InterCode utilizes a tree-structured search methodology, requiring merely logarithmic space and time. This approach is notably beneficial for expansive search spaces or environments prone to frequent updates (e.g., Docker environments, customized systems). By integrating online RL training into InterCode, our proof-of-concept environment aims to create code interpreter designed for large code repositories.

During interaction with \aecopilot, each step the agent receives $-1$ reward as the cost of time.
After $15$ steps or the agent explicitly terminates, if the correct file is identified, a $+15$ reward is given; otherwise a $-15$ reward is given.





\paragraph{\taxi.} We extended the OpenAI Gym's Taxi environment to introduce a higher level of challenge, thereby creating the ``\taxi" environment. This game concludes prematurely if the player commits any invalid action, such as colliding with a wall, or incorrectly picking up or dropping off passengers at unauthorized locations. This modification crucially elevates the task's difficulty by eliminating the opportunity for the model to correct its mistakes after a wrong decision—a common allowance in the standard environment. 

We applied curriculum learning to \taxi.
In the designed pickup curriculum, we assign a positive reward $20$ and terminate the environment  after the driver successfully pickup the passenger.
In the dropoff stage, the pickup reward is retained, but the driver needs to further dropoff the passenger at destination to receive the full reward.

\paragraph{\alfw.}
Our study leverages ALFWorld \citep{cote2019textworld,shridhar2020alfworld}, a multi-turn platform tailored for simulating household tasks by converting the graphical representation of a house into descriptive language.
A robot in is required to complete certain tasks based on the descriptions.
This benchmark has gained recognition to evaluate LLM agents, with studies like \citet{arabzadeh2024better} demonstrating its efficacy. 
Our focus on the tomato picking task stems from its optimal mix of simplicity and representativeness. 


\section{Experimental Results} \label{sec:exp_results}

\input{fig_tex/taxi_plot}


\rebut{To verify our approach, we apply \rlft on GPT-2 XL \citep{Radford2019LanguageMA}.}
\Cref{table:all-rst} presents a comprehensive evaluation of various models' performance across different environments.
LMs still face challenges in multi-step decision-making in interactive environments, and \rlft has significantly improved their decision-making capabilities in complex environments.
This method not only utilizes the inherent strengths of LMs in \emph{reflection} but also closely aligns with the multi-step decision-making process intrinsic to RL. Our findings highlight the potential of merging advanced prompting techniques with LMs to address complex RL tasks, establishing a new benchmark for future research in this field. 

\paragraph{Open source models and commercial GPT models.}
We evaluated three open-source 7B models with necessary prompt engineering such as ReAct and memory mechanism included.
These models all perform poorly on the three tasks, except for Mistral 7B on \autoexplore depth 1.
We also examined GPT-3.5-turbo and GPT-4 (version 1106) through Azure OpenAI API.
\rebut{GPT-4 can achieve a success rate of 71\% in \autoexplore depth 1, 81\% in depth 2, and 84\% in \alfw; meanwhile, GPT-3.5-turbo achieves a success rate of 31\% in \autoexplore depth 1, 8\%  in depth 2, and 6\% in \alfw. During the evaluation, we noticed potential data contamination of these two models: GPT-4 can sometimes identify near-optimal actions without extensive exploration of the space. In the \taxi environment, the success rates of the dropoff curriculum for GPT-4 and GPT-3.5-turbo are both 0\%. Even though GPT-4 has 70\% chance executing a valid action in each step, it is prone to failure upon committing minor errors along the long navigation path during multi-turn interactions.}
These observations suggest that even powerful LLMs may still need online RL training for multi-turn interactions.



\paragraph{SFT is not enough.}
Supervised fine-tuning (SFT) has been widely used offline to improve LMs' performance on specific tasks. However, our results (\Cref{table:all-rst}) indicate that SFT alone is not sufficient for complex RL tasks requiring multi-step decision-making. While SFT enhances task-specific knowledge, it fails to solve problems requiring deep reasoning, planning, and reflection.

\paragraph{Reflection helps learning.}
Incorporating reflective processes into LLMs significantly enhances decision-making and learning from past actions. Our comparative analysis between models with and without reflection capabilities highlights the importance of reflection for advanced understanding and adaptability in RL tasks. 
As shown in \Cref{fig:taxi_conv}, the curves representing online RL without reflection are constantly below the curve of \rlft.
\Cref{fig:taxi_curriculum} shows a similar result.



\paragraph{Reflecting from mistakes is beneficial.}
The philosophy of ``learning from mistakes" plays a meaningful role in \rlft.
Without negative reflection samples, the model’s performance would be worse (in absolute difference) than the model trained with both positive and negative data.
For \autoexplore, the test accuracies without negative examples are $33\%$ and $12\%$ for each curriculum, compared with $36\%$ and $17\%$ with negative examples.
As shown in \Cref{fig:neg_reflect}, the solid curve represents the integration of negative examples into the SFT dataset, and we observed a faster convergence during RLFT.

\input{fig_tex/ae_plot}

\paragraph{Curriculum learning (CL) accelerates learning.}
As shown in the top two curves in \Cref{fig:taxi_curriculum}, CL accelerates the learning curve for complex RL tasks by structuring the training process with challenging tasks. To ensure a fair evaluation, both learning approaches (\rlft with and without CL) are pre-trained with the same reflection dataset during the SFT phase. The curriculum learning approach begins with an initial RL training phase focused on the pickup curriculum, followed by the dropoff curriculum. Without curriculum learning, the model is trained directly using the dropoff curriculum, resulting in slightly inferior performance.

\input{fig_tex/curriculum.tex}

\paragraph{Sensitivity of the policy model with respect to the reflection model.}
\rebut{For \taxi pickup subtask, using the same policy model after \rlft, we switch the reflection model to GPT-2 Small 0.12B and Mistral 7B SFTed with the reflection data.
The results for GPT-2 Small 0.12B, GPT-2 XL 1.56B, and Mistral 7B are $55\%$, $58\%$ (as in \Cref{table:all-rst}), and $64\%$.
This phenomenon indicates that the policy model is not extremely sensitive to the robustness/accuracy of the reflection model as the policy model can easily adapt.
Additionally, using a more capable reflection model can improve the performance.}

%% file: fig_tex/rst_table.tex
\begin{table*}[ht]
\centering
\begin{tabular}{c|l|cc|cc|c}
\specialrule{.15em}{.05em}{.05em}
& \multirow{2}{*}{Model} & \multicolumn{2}{c|}{\autoexplore} 
& \multicolumn{2}{c|}{\taxi}  & \multirow{2}{*}{\alfw} \\ 
&    & Depth 1 &  Depth 2 & Pickup & $+$Dropoff$^\star$ & \\
\specialrule{.1em}{.05em}{.05em}
\specialrule{.1em}{.05em}{.05em}
 \multirow{3}{*}{Open Source}
& Mistral 7B & 34\% & 3\% & 7\% & 0\%  & 0\% \\
& Llama2 7B-chat &  2\%  & 1\%  & 3\% & 0\% & 0\% \\
& Orca-2 7B & 6\% & 1\%  & 1\% & 0\%  & 0\% \\
\specialrule{1pt}{1pt}{1pt}
SFT Only & GPT-2 XL 1.56B & 4\% & 9\% & 7\%  & 0\%  & 0\% \\
\midrule
\rebut{RLFT Only} & GPT-2 XL 1.56B & 12\% & 3\% & 2\%  & 0\%  & 0\% \\
\midrule
\rebut{SFT$+$RLFT (w/o reflection)} & GPT-2 XL 1.56B & 20\% & 4\% & 6\%  & 0\%  & 66\% \\
\midrule
\rebut{SFT$+$RLFT (w/o negative)} & GPT-2 XL 1.56B & 33\% & 12\% & -  & - & - \\
\specialrule{1pt}{1pt}{1pt}
\rlft (Ours) & GPT-2 XL 1.56B & {\bf 36\%} & {\bf 17\%} &  {\bf 58\%} & {\bf 29\%}  & {\bf 74\%} \\
\specialrule{.15em}{.05em}{.05em}
\end{tabular}
\caption{Testing performance (average success rate) of open source models \citep{jiang2023mistral,touvron2023llama,mitra2023orca}, \rebut{GPT-2 XL fine-tuned with} baselines, and with \rlft.
ReAct and memory mechanism, as shown in \Cref{fig:rlft}, have been incorporated to improve performance.
\rebut{For conciseness, we have not performed prompt optimization for the open-source models, and their performance could potentially be improved with different prompting techniques in the future.}
{\bf Explanation for baselines:}
``SFT$+$RL (w/o reflection)'' means the policy model is the only model involved, and the reflection field is removed from SFT data.
``SFT$+$RL (w/o negative)'' means there are no negative examples in SFT data, so both the reflection model and the policy model are trained on expert demonstrations.
We only ran this ablation on \autoexplore.
{\bf Explanation for tasks:}
For \autoexplore, we tested on $44$ user queries, each with $10$ runs. ``Depth $i$'' includes the tasks with target file depth exactly $i$.
For \taxi, we ran on $100$ random maps. ``Pickup'' computes the success rate of picking up the passenger, and ``$+$Dropoff'' computes the overall success rate.
For \alfw, we tested on $4$ tasks, each with $25$ runs.
}
\label{table:all-rst}
\end{table*}

%% file: fig_tex/taxi_plot.tex
\begin{figure}[t]
    \centering
	\pgfplotstableread[col sep=comma]{fig_tex/taxi_convergence.csv}{\datatable}
    \tikzsetnextfilename{baseline-compare-large}
	\begin{tikzpicture}
	\begin{axis}[
    	axis x line*=bottom,
        ylabel={Success Rate (\%)},     
        xlabel={Iters},     
    	tick label style={font=\small}, 
        ymin=0, ymax=85,
        xmin=0, xmax=5000,
    	width=0.48\textwidth,
    	height=3.5cm,
	 legend style={/tikz/every even column/.append style={column sep=0.1cm}, font = \tiny, at={(0.5,1.1)}, anchor=north,legend columns=3, draw=none, fill=none},
	]
\addplot [draw=msblue, line width=1pt, line join=round, line join=round] table[x expr=\thisrow{step}, y expr=\thisrow{217_rl_finetune} * 100.0] {\datatable};

\addplot [draw=msgreen, dashed, line width=1pt, line join=round, line join=round] table[x expr=\thisrow{step}, y expr=\thisrow{227_rl_finetune} * 100.0] {\datatable};

\addplot [draw=msred, dashed, line width=1pt, line join=round, line join=round] table[x expr=\thisrow{step}, y expr=\thisrow{225_rl_finetune} * 100.0] {\datatable};

    
	\legend{\rlft, SFT + RL, RL}
	\end{axis}

	\end{tikzpicture}

\caption{Training success rates of different training methods with GPT-2 XL in the pickup curriculum of the \taxi environment. We compared different RL methods for 5000 iterations during RLFT. SFT with 5000 iterations would only achieve 7\% success rate, hence only RL methods are shown.
}
    \label{fig:taxi_conv}
\end{figure}

%% file: fig_tex/ae_plot.tex
\begin{figure}[H]
    \centering
	\pgfplotstableread[col sep=comma]{fig_tex/ae_rst.csv}{\datatable}
    \tikzsetnextfilename{curriculum-compare}
	\begin{tikzpicture}
	\begin{axis}[
    	axis x line*=bottom,
        ylabel={Success Rate (\%)},     
        xlabel={Iters},     
    	tick label style={font=\small}, 
        ymin=20, ymax=70,
        xmin=0, xmax=5000,
    	width=0.48\textwidth,
    	height=3.5cm,
	 legend style={/tikz/every even column/.append style={column sep=0.1cm}, font = \tiny, at={(0.5,1.1)}, anchor=north,legend columns=1, draw=none, fill=none},
	]

\addplot [draw=msred, line width=1pt, line join=round, line join=round] table[x expr=\thisrow{step}, y expr=\thisrow{zhou_536} * 100.0] {\datatable};

\addplot [draw=msblue, line width=1pt, line join=round, dashed, dashed] table[x expr=\thisrow{step}, y expr=\thisrow{518_rl_finetune} * 100.0] {\datatable};



	\legend{Positive + Negative Examples, Positive Examples Only}
	\end{axis}

	\end{tikzpicture}
\caption{Training success rate with and without negative examples in the \autoexplore setting, each assessed in a single run. When negative examples are excluded, the training process exhibits decreased speed and lacks smoothness.} \label{fig:neg_reflect}
\end{figure}

%% file: fig_tex/curriculum.tex
\begin{figure}[H]
    \centering
	\pgfplotstableread[col sep=comma]{fig_tex/taxi_curriculum.csv}{\datatable}
    \tikzsetnextfilename{curriculum-compare}
	\begin{tikzpicture}
	\begin{axis}[
    	axis x line*=bottom,
        ylabel={Success Rate (\%)},     
        xlabel={Iters},     
    	tick label style={font=\small}, 
        ymin=0, ymax=50,
        xmin=0, xmax=5000,
    	width=0.48\textwidth,
    	height=3.5cm,
	 legend style={/tikz/every even column/.append style={column sep=0.1cm}, font = \tiny, at={(0.5,1.1)}, anchor=north,legend columns=2, draw=none, fill=none},
	]

\addplot [draw=msred, line width=1pt, line join=round, line join=round] table[x expr=\thisrow{step}, y expr=\thisrow{382_rl_finetune} * 100.0] {\datatable};

\addplot [draw=msblue, line width=1pt, line join=round, dashed] table[x expr=\thisrow{step}, y expr=\thisrow{384_rl_finetune} * 100.0] {\datatable};

\addplot [draw=orange, line width=1pt, line join=round, dashed] table[x expr=\thisrow{step}, y expr=\thisrow{521_rl_finetune} * 100.0] {\datatable};

\addplot [draw=msgreen, line width=1pt, line join=round, dashed] table[x expr=\thisrow{step}, y expr=\thisrow{522_rl_finetune} * 100.0] {\datatable};

	\legend{\rlft (w/ CL), \rlft (w/o CL), RL + SFT, RL}
	\end{axis}

	\end{tikzpicture}

\caption{Comparison of training success rates in the drop-off curriculum in the \taxi environment. The top two curves represent \rlft; ``w/ CL'' means the experiment incorporates curriculum learning (CL) and is trained with the pickup curriculum. The bottom two dashed curves represent online RL without reflection. All single run.}

 \label{fig:taxi_curriculum}
\end{figure}

%% file: 6_conclusion.tex
\section{Discussion and Conclusion} \label{sec:disc_conc}



\paragraph{Risk, impact, and responsible AI.}
In this study, we adhere to principles of Responsible AI by ensuring transparency, efficiency, and security in both the training and evaluation stages. An exemplar of our commitment is the development of \aesandbox, designed to reduce the risk of security issues in the file system.
Recognizing the importance of ethical considerations and the social impact of our work, we pledge to engage in continuous evaluation of LMs's performance in multi-step environment. 

\paragraph{Limitations.} 
Our study, while comprehensive, acknowledges certain limitations.
Although \alfw benchmark is multimodal, this study primarily focued on the text representation, leaving the examination of multimodal models and cross-attention encoding of other modalities (such as images and audio) for future work.
Comparisons with commercial models is discussed in \Cref{sec:exp_results}, but the proprietary nature and potential biases (e.g., unknown training data) limit a fair comparison with open-source models. Standardized benchmarks in the field are needed for further evaluation. 
Lastly, the reflection data utilized in our study is generated by GPT-4, which may not fully capture the distribution of real human data.
This indicates the importance of integrating more authentic human-generated data in future evaluations.

\paragraph{Future direction.} 
The primary goal of this study is to create an efficient online RL pipeline for LMs to perform multi-step problem solving. Building on this foundation, future research directions may explore the scalability of \rlft to develop larger foundation models, enabling them to adapt to previously unseen environments with out-of-domain generalization capabilities. 
The two-player design in our framework may naturally be extended to other multi-agent settings where language models can show their strengths.
Another future direction is to train the reflection model in RLFT stage as we freeze it because of the interference with the policy model (\Cref{sec:same_model}), which will improve the reasoning ability of language models for decision-making tasks.



%% file: 9_appendix.tex
\input{9_appendix_ppo}

\input{9_appendix_pipeline}
\input{9_appendix_ae}

\input{9_appendix_exp}

\input{9_appendix_qualitative_obs}

%% file: 9_appendix_ppo.tex
\section{Discussion on PPO} \label{sec:ppo}

Proximal policy optimization (PPO) is an advanced policy gradient method, which aims to take the largest possible improvement step on a policy while ensuring the deviation from the previous policy is reasonably small.
The update step is
\begin{align*}
    \theta_{t + 1} = \arg \max_\theta  \bbE_{s, a} \left[\rule{0cm}{0.5cm}\right. \min \left\{\rule{0cm}{0.5cm}\right. \frac{\pi_\theta (a | s)}{\pi_{\theta_t} (a | s)} A^{\pi_{\theta_t}} (s, a), \\
    \clip \left(\frac{\pi_\theta (a | s)}{\pi_{\theta_t} (a | s)}, 1 - \epsilon, 1 + \epsilon \right) A^{\pi_{\theta_t}} (s, a) \left.\rule{0cm}{0.5cm} \right\} \left.\rule{0cm}{0.5cm}\right],
\end{align*}
where $\epsilon$ usually takes small values such as $0.1$ or $0.2$.

In practice, we found that PPO did not work well.
For tasks with a large state space, action space, and a long horizon, the training processes were constantly unstable, with sudden drops of the expected total reward.
Such tasks pose high difficulty for the value function estimator to learn the value functions when the policy network changes.
Most importantly, inherent randomness (including dropout, padding length in different batches, \texttt{top\_p} and \texttt{top\_k}) of LMs results in high sensitivity for the terms of $\frac{\pi_\theta (a | s)}{\pi_{\theta_t} (a | s)}$.
Though we want the policy to sample actions with small possibilities (e.g., $\pi_{\theta_t} (a | s) < \varepsilon$) to encourage exploration, high sensitivity will result in such values becoming to $0$ in almost all the future re-evaluations.

%% file: 9_appendix_pipeline.tex
\section{Illustrations of Pipeline} \label{sec:illustration_pipeline}

\rebut{In this section we present detailed versions of \Cref{fig:rlft}.
\Cref{fig:rlft_data_gen} is the illustration for data generation.
\Cref{fig:rlft_sft} is the illustration for SFT stage.
\Cref{fig:rlft_rlft} is the illustration for RLFT stage.}

\input{fig_tex/system_data_gen}

\input{fig_tex/system_sft}

\input{fig_tex/system_rlft}

%% file: fig_tex/system_data_gen.tex
\begin{figure*}
    \centering
\begin{tikzpicture}[node distance=2cm, auto,
block/.style={
  rectangle,
  draw,
  text width=2.5cm,
  text centered,
  rounded corners,
  minimum height=2.5cm,
},
line/.style={
  draw, 
  -latex',
  shorten >=2pt,
  line width=0.3mm,
},
train_line/.style={
  draw=none, 
  -latex',
  shorten >=2pt,
  line width=0.4mm,
  dashed,
  dash pattern=on 6pt off 3pt on 1pt off 3pt,
  text=black, 
  opacity=0 
},
cloud/.style={
  draw,
  ellipse,
  minimum height=2em
}]

\tikzstyle{fitbox} = [rectangle, dashed, draw=black, inner sep=0.5cm]

\node [block, draw=none] (env) {\includegraphics[width=1.8cm]{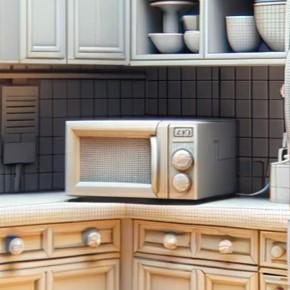} \\ Environment};

\node [block, right=0.5cm of env, yshift=1.5cm] (obs) {Observation\\
\vspace{0.2cm}
  \parbox{1.8cm}{ 
  \centering
  \tiny  You are in the middle of a room. You see a cabinet 18, a cabinet 3, a ...}};

\node [block, right=0.5cm of env, yshift=-1.5cm] (actions) {Possible Actions
\\~\\
\vspace{0.2cm}
  \parbox{1.8cm}{ 
  \centering
  \tiny  1. Close microwave\\2. Put tomato in microwave\\3. Go to cabinet\\ ...}
  
};

\node [block, circle, minimum height=1cm, text width=1.5cm, right=0.55cm of obs, yshift=0cm, draw=none, fill=cyan!20] (reflect-agent) {\includegraphics[width=0.8cm]{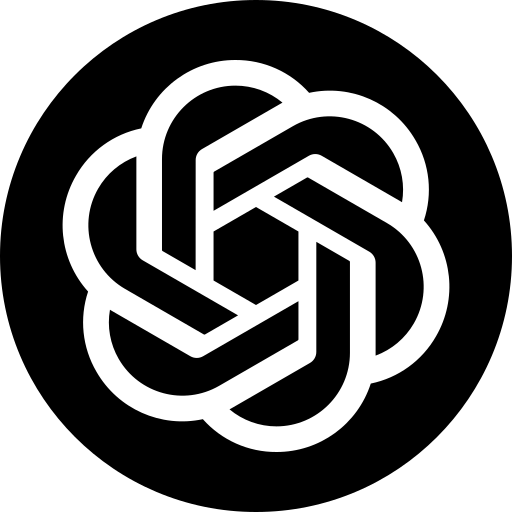}\\ GPT-4 Reflector};

\node [block, right=0.5cm of reflect-agent, fill=yellow!30] (reflection) {Reflection \\
  \centering
  \vspace{0.2cm}
  \parbox{2.3cm}{ 
    \tiny 
    \setlength{\baselineskip}{8pt} 
    Taking the tomato to the microwave is correct. Next step is to place the tomato ...
  }
};
\node [block, right=0.49cm of actions, fill=blue!20] (memory) {\includegraphics[width=0.5cm]{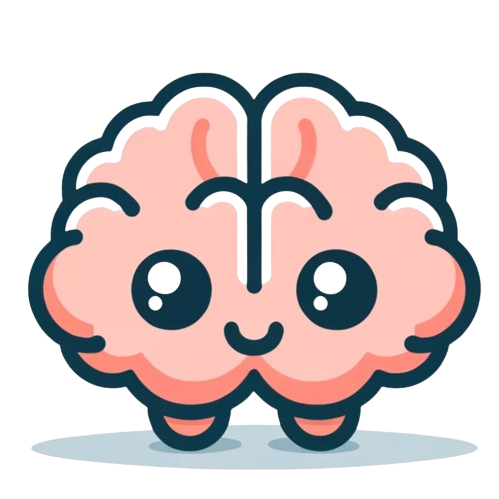} \textbf{Memory}\\
  \vspace{0.2cm}
  \parbox{2.5cm}{ 
  \centering
  \tiny \textbf{\tiny \textcolor{red}{Task: Cook Tomato}}\\
  Act 0: Go to fridge\\
  Obs 0: It is closed \\
  ...\\
  Act 4: Open microwave\\
  Obs 4: You see a mug ...
  }
};

\node [block, right=0.2cm of reflection, draw=none] (database) {\includegraphics[width=1.9cm]{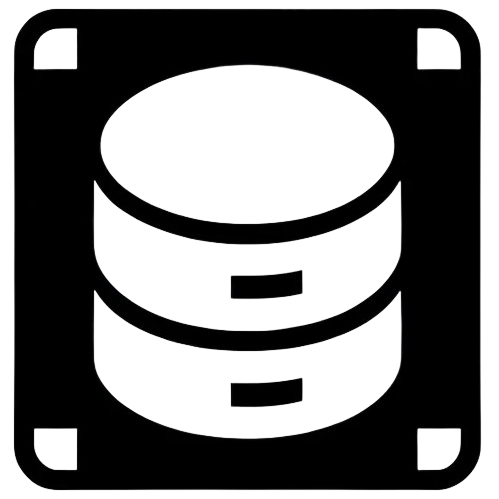}\\Database};

\node [block, circle, minimum height=1cm, text width=1.5cm, right=0.5cm of memory, draw=none, fill=orange!20] (policy-agent) {\includegraphics[width=0.8cm]{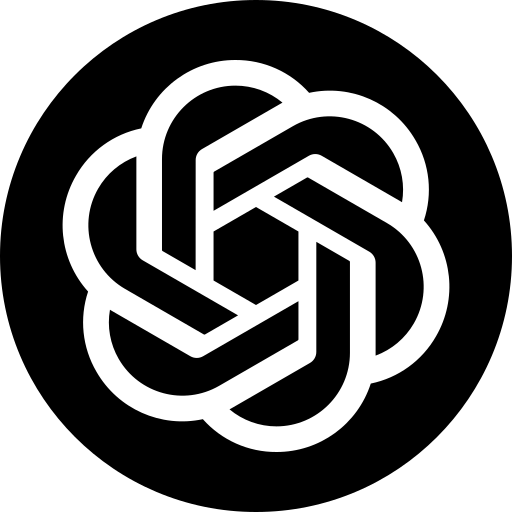} \\ GPT-4 Policy};

\node [right=0.58cm of policy-agent, fill=green!20, rounded corners] (policy-action) {\includegraphics[width=0.3cm]{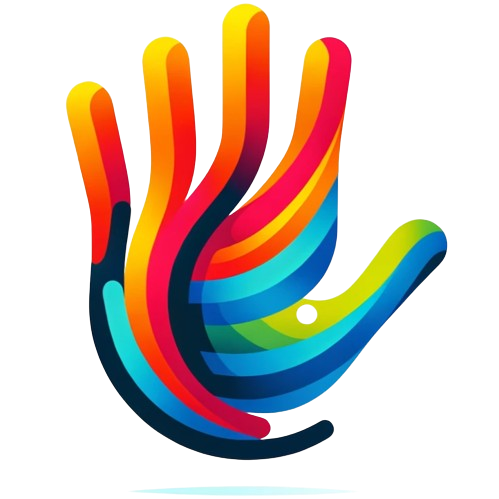} Action: 2};

\path [line] (env) -- (obs);
\path [line] (env) -- (actions);
\path [line, <->] (memory) -- (reflect-agent);
\path [line] (obs) -- (reflect-agent);
\path [line] (reflect-agent) -- (reflection);
\path [line] (reflection) -- (policy-agent);
\path [line] (actions.south) -- ++(0, -0.5) -| (policy-agent.south);

\path [line] (policy-agent.east) -- (policy-action);
\path [line] (policy-action.east) -- ++(0.3, 0) -- ++ (0, 4.5) -| (env.north);
\path [line] (policy-action) --  (database.south);

\path [line,shorten >=-4mm] (reflection) -- (database);

\path[train_line,->] (policy-agent) edge [loop below, looseness=20, out=-55, in=-35, distance=1cm] node [right, opacity=0] {\parbox{4cm}{Policy Gradient\\$Q(s,a) \nabla_\theta \ln \pi_\theta (a|s)$}} (policy-agent); 

\end{tikzpicture}
   \caption{Pipeline of \rlft data generation.} \label{fig:rlft_data_gen}
\end{figure*}

%% file: fig_tex/system_sft.tex
\begin{figure*}
    \centering
\begin{tikzpicture}[node distance=2cm, auto,
block/.style={
  rectangle,
  draw,
  text width=2.5cm,
  text centered,
  rounded corners,
  minimum height=2.5cm,
},
line/.style={
  draw, 
  -latex',
  shorten >=2pt,
  line width=0.3mm,
},
train_line/.style={
  draw, red,
  -latex',
  shorten >=2pt,
  line width=0.4mm,
  dashed,
  dash pattern=on 6pt off 3pt on 1pt off 3pt,
  text=black
},
cloud/.style={
  draw,
  ellipse,
  minimum height=2em
}]

\tikzstyle{fitbox} = [rectangle, dashed, draw=black, inner sep=0.5cm]

\node [block, draw=none] (env) {};

\node [block, right=0.5cm of env, yshift=1.5cm] (obs) {Observation\\
\vspace{0.2cm}
  \parbox{1.8cm}{ 
  \centering
  \tiny  You are in the middle of a room. You see a cabinet 18, a cabinet 3, a ...}};

\node [block, right=0.5cm of env, yshift=-1.5cm] (actions) {Possible Actions
\\~\\
\vspace{0.2cm}
  \parbox{1.8cm}{ 
  \centering
  \tiny  1. Close microwave\\2. Put tomato in microwave\\3. Go to cabinet\\ ...}
  
};

\node [block, circle, minimum height=1cm, text width=1.5cm, right=0.5cm of obs, yshift=0cm, draw=none, fill=cyan!20] (reflect-agent) {\includegraphics[width=0.8cm]{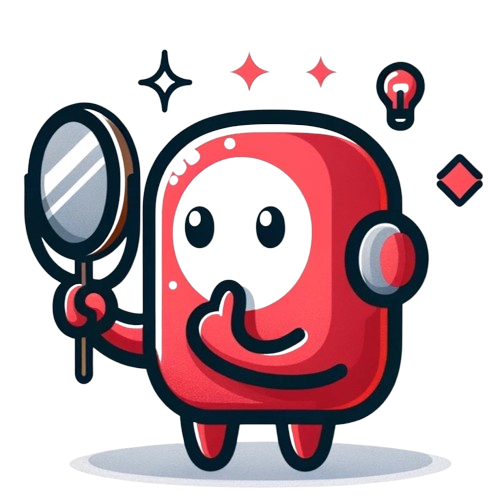}\\Reflect Agent $\wh{R}_\phi$};

\node [block, right=0.5cm of reflect-agent, fill=yellow!30] (reflection) {Reflection \\
  \centering
  \vspace{0.2cm}
  \parbox{2.3cm}{ 
    \tiny 
    \setlength{\baselineskip}{8pt} 
    Taking the tomato to the microwave is correct. Next step is to place the tomato ...
  }
};
\node [block, right=0.49cm of actions, fill=blue!20] (memory) {\includegraphics[width=0.5cm]{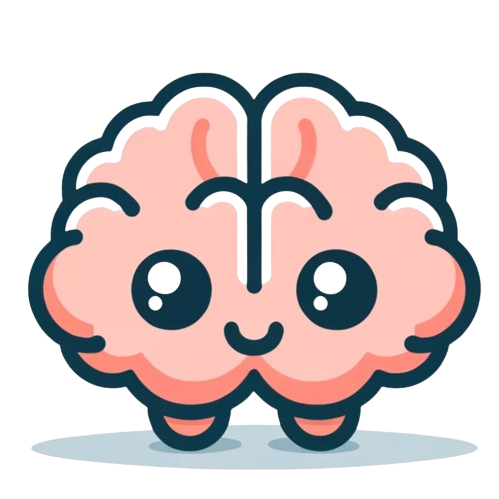} \textbf{Memory}\\
  \vspace{0.2cm}
  \parbox{2.5cm}{ 
  \centering
  \tiny \textbf{\tiny \textcolor{red}{Task: Cook Tomato}}\\
  Act 0: Go to fridge\\
  Obs 0: It is closed \\
  ...\\
  Act 4: Open microwave\\
  Obs 4: You see a mug ...
  }
};

\node [block, right=0.2cm of reflection, draw=none] (database) {\includegraphics[width=1.9cm]{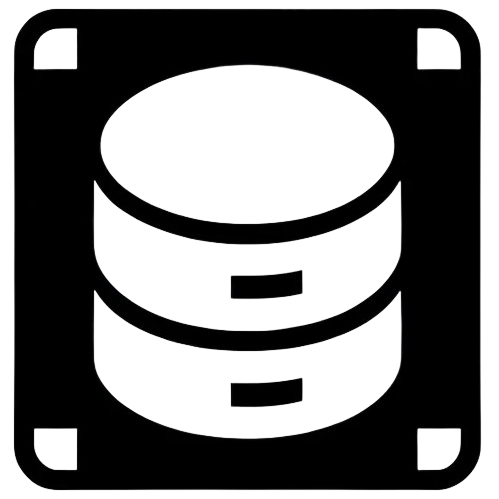}\\Database};

\node [block, circle,  minimum height=1cm, text width=1.5cm, right=0.5cm of memory, draw=none, fill=orange!20] (policy-agent) {\includegraphics[width=0.8cm]{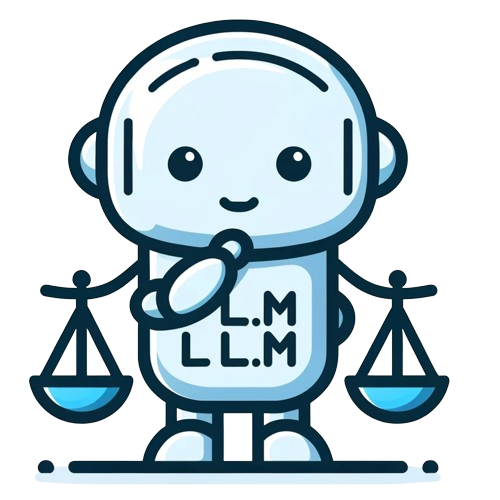} \\ Policy Agent $\pi_\theta$};

\node [right=0.58cm of policy-agent, fill=green!20, rounded corners] (policy-action) {\includegraphics[width=0.3cm]{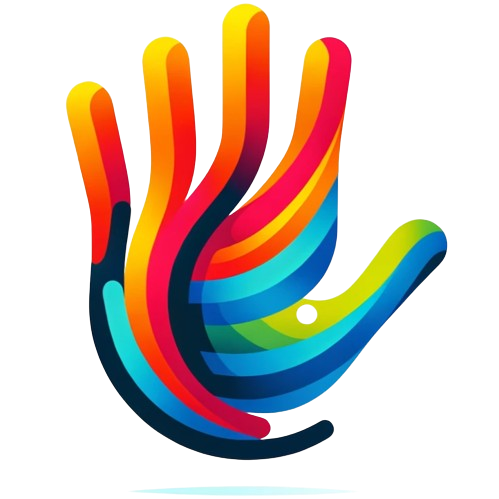} Action: 2};

\path [line] (memory) -- (reflect-agent);
\path [line] (obs) -- (reflect-agent);
\path [train_line] (reflection) -- (reflect-agent);
\path [train_line] (reflection) -- (policy-agent);
\path [line] (actions.south) -- ++(0, -0.5) -| (policy-agent.south);

\path [train_line] (policy-action) -- (policy-agent.east);
\path [train_line] (database.south) -- (policy-action);

\path [train_line,shorten <=4.5mm] (database) -- (reflection);

\path[train_line,->] (reflect-agent) edge [loop above, looseness=20, out=80, in=100, distance=1cm] node [right] {$\cL_{\textup{reflect}} (\phi)$} (reflect-agent);

\path[train_line,->] (policy-agent) edge [loop below, looseness=20, out=-55, in=-35, distance=1cm] node [right] {$\cL_{\textup{policy}} (\theta)$} (policy-agent);

\path[train_line,->,opacity=0] (policy-agent) edge [loop below, looseness=20, out=-55, in=-35, distance=1cm] node [right, opacity=0] {\parbox{4cm}{Policy Gradient\\$Q(s,a) \nabla_\theta \ln \pi_\theta (a|s)$}} (policy-agent); 

\end{tikzpicture}

   \caption{Pipeline of \rlft SFT stage.
   }
    \label{fig:rlft_sft}
\end{figure*}

%% file: fig_tex/system_rlft.tex
\begin{figure*}
    \centering
\begin{tikzpicture}[node distance=2cm, auto,
block/.style={
  rectangle,
  draw,
  text width=2.5cm,
  text centered,
  rounded corners,
  minimum height=2.5cm,
},
line/.style={
  draw, 
  -latex',
  shorten >=2pt,
  line width=0.3mm,
},
train_line/.style={
  draw, red,
  -latex',
  shorten >=2pt,
  line width=0.4mm,
  dashed,
  dash pattern=on 6pt off 3pt on 1pt off 3pt,
  text=black
},
cloud/.style={
  draw,
  ellipse,
  minimum height=2em
}]

\tikzstyle{fitbox} = [rectangle, dashed, draw=black, inner sep=0.5cm]

\node [block, draw=none] (env) {\includegraphics[width=1.8cm]{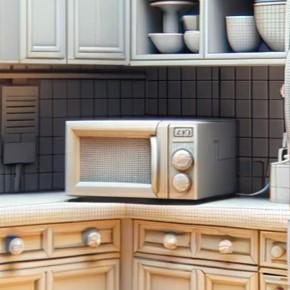} \\ Environment};

\node [block, right=0.5cm of env, yshift=1.5cm] (obs) {Observation\\
\vspace{0.2cm}
  \parbox{1.8cm}{ 
  \centering
  \tiny  You are in the middle of a room. You see a cabinet 18, a cabinet 3, a ...}};

\node [block, right=0.5cm of env, yshift=-1.5cm] (actions) {Possible Actions
\\~\\
\vspace{0.2cm}
  \parbox{1.8cm}{ 
  \centering
  \tiny  1. Close microwave\\2. Put tomato in microwave\\3. Go to cabinet\\ ...}
  
};

\node [block, circle, minimum height=1cm, text width=1.5cm, right=0.5cm of obs, yshift=0cm, draw=none, fill=cyan!20] (reflect-agent) {\includegraphics[width=0.8cm]{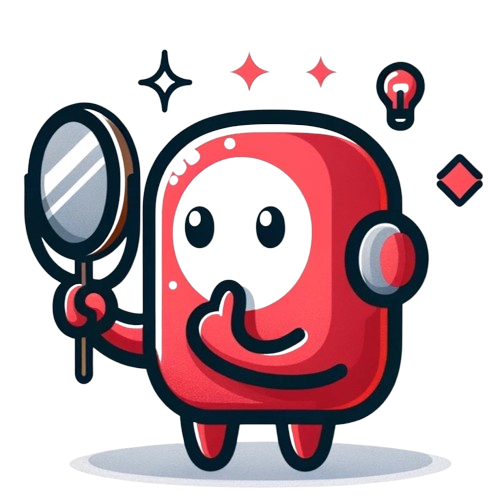}\\Reflect Agent $\wh{R}$};

\node [block, right=0.5cm of reflect-agent, fill=yellow!30] (reflection) {Reflection \\
  \centering
  \vspace{0.2cm}
  \parbox{2.3cm}{ 
    \tiny 
    \setlength{\baselineskip}{8pt} 
    Taking the tomato to the microwave is correct. Next step is to place the tomato ...
  }
};
\node [block, right=0.49cm of actions, fill=blue!20] (memory) {\includegraphics[width=0.5cm]{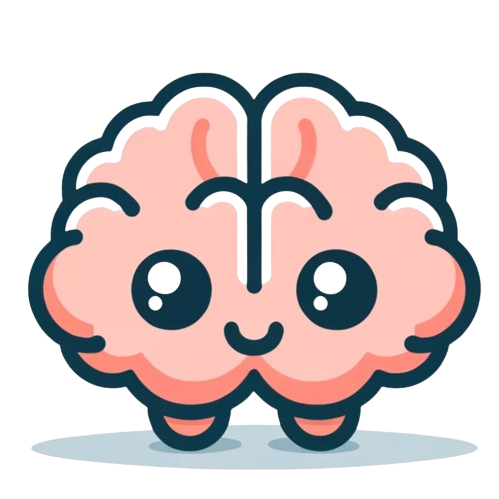} \textbf{Memory}\\
  \vspace{0.2cm}
  \parbox{2.5cm}{ 
  \centering
  \tiny \textbf{\tiny \textcolor{red}{Task: Cook Tomato}}\\
  Act 0: Go to fridge\\
  Obs 0: It is closed \\
  ...\\
  Act 4: Open microwave\\
  Obs 4: You see a mug ...
  }
};

\node [block, circle,  minimum height=1cm, text width=1.5cm, right=0.5cm of memory, draw=none, fill=orange!20] (policy-agent) {\includegraphics[width=0.8cm]{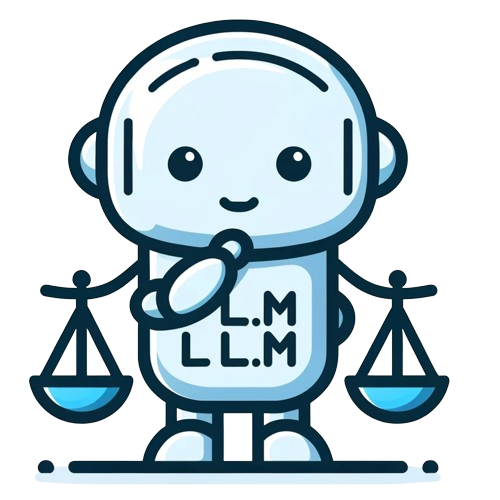} \\ Policy Agent $\pi_\theta$};

\node [right=0.58cm of policy-agent, fill=green!20, rounded corners] (policy-action) {\includegraphics[width=0.3cm]{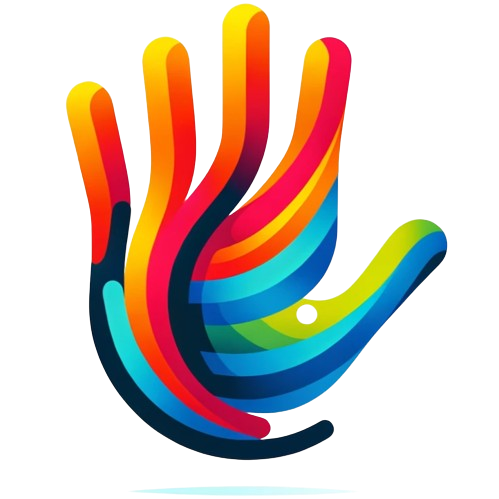} Action: 2};

\path [line] (env) -- (obs);
\path [line] (env) -- (actions);
\path [line, <->] (memory) -- (reflect-agent);
\path [line] (obs) -- (reflect-agent);
\path [line] (reflect-agent) -- (reflection);
\path [line] (reflection) -- (policy-agent);
\path [line] (actions.south) -- ++(0, -0.5) -| (policy-agent.south);

\path [line] (policy-agent.east) -- (policy-action);
\path [line] (policy-action.north) -- ++ (0, 4.2) -| (env.north);

\path [train_line] (env.south) -- ++(0, -2.5) -- ++ (1.0, 0.0) -| ([xshift=0.3cm]policy-agent.south) 
node[midway, above, rounded corners, xshift=-8.5cm, yshift=0.1cm] {[ Reward: +1 ]};

\path[train_line,->] (policy-agent) edge [loop below, looseness=20, out=-55, in=-35, distance=1cm] node [right] {\parbox{4cm}{Policy Gradient\\$Q(s,a) \nabla_\theta \ln \pi_\theta (a|s)$}} (policy-agent);

\end{tikzpicture}

   \caption{Pipeline of \rlft RLFT stage.}
    \label{fig:rlft_rlft}
\end{figure*}

%% file: 9_appendix_ae.tex
\section{Autonomous Exploration Details}
\label{sec:ae_details}

Autonomous exploration in a well-organized repository can reduce the number of reads of files to a large extent.
Ideally, if the repository has $n$ files and is organized as a $k$-ary tree, the best language model only takes $O(k + \log_k n)$ (compared to $O(n)$ using exhaustive enumeration) commands to identify the correct file, then proceed with the specific needs of reading, editing, and executing.
This serves as an motivation of autonomous exploration benchmark.

\subsection{Autonomous Exploration Sandbox}

\aesandbox is a sandbox protecting the original repository from modification.
An instance of \aesandbox could be initialized with the path to the original repository, then this instance will create a temporary directory in a specified location (could possibly be a ram disk) and make a duplication of the original repository.
\aesandbox supports two main functions:
\begin{enumerate}
    \item {\bf Executing system commands:} For the purpose of our work, commands such as ``\texttt{cd}'', ``\texttt{ls}'', ``\texttt{cat}'', ``\texttt{head}'', ``\texttt{tail}'', ``\texttt{echo}'', ``\texttt{python}'' and ``\texttt{pip}'' are supported to enable document retrieval and coding.

    \item {\bf Tracking changed files:} The user of \aesandbox could call a function to get the list of the changed files and their contents compared to the original status when creating the sandbox.
\end{enumerate}

\subsection{Autonomous Exploration Copilot}

\aecopilot is an agent mediating between language models, humans, and \aesandbox.
An instance of \aecopilot could be initialized with a natural language question and the corresponding repository to work in.
The main function of \aecopilot is to give natural language descriptions of the current autonomous exploration task for either human or language models to make decisions.
The interaction proceeds in loops ($k$ starts from $0$):
\begin{itemize}
    \item {\bf Step $3 k + 1$: Prompting.} \aecopilot compiles a prompt $p_k$ given the current status of \aesandbox, which includes the question, current working directory (cwd) in the repository, files and folders under cwd (optional, can be used to replace \texttt{ls} and reduce interaction), historical commands $c_0, c_1, \ldots, c_{k - 1}$ from the human or language model, and execution result of the last command $c_{k - 1}$.

    \item {\bf Step $3 k + 2$: Querying.} \aecopilot sends the prompt $p_k$ to human or LM and gets the response.
    This response may contain excessive information such as analysis of the current situation (which is a typical behavior of GPT-4), so \aecopilot needs to extract system command $c_k$ from the response.

    \item {\bf Step $3 k + 3$: Executing.} \aecopilot sends the system command $c_k$ to \aesandbox and gets the execution results.
    The results contain standard output and standard error, such as the file content after ``\texttt{cat}'' and runtime error of ``\texttt{python}''.
\end{itemize}
The interaction ends when the response in step $3 k + 2$ contains an exit signal stipulated in the prompt.

\aecopilot is capable of prompting GPT-4 to do the entire task, while for the smaller models in this work we only set the goal to be a subtask (file identification).

\subsection{Labeled Dataset} \label{sec:dataset}

The licenses are bounded by each open-source repository used in this dataset.

Using GPT-4 from Azure OpenAI service, we constructed a synthetic repository called ``Coffee Company'', which contains documents (in \texttt{.md} format), codes (in various programming languages), and database files (in \texttt{.csv} format).
This repository contains around 12 million tokens.
In addition, we downloaded 12 open-source repositories containing codes and documentations from GitHub.

After collecting the repositories, we built a labeled dataset regarding autonomous exploration.
Each datum in the dataset contains the following fields: the name of the repository $n$, a natural language question $q$, an answer to this question $a$, the related file $f$, and the shortest system command path to reach this file $c^\star = (c^\star_0, c^\star_1, \ldots, c^\star_{L-1})$.
As a start point, this work focus on an important step in autonomous exploration: find the correct file $f$ given the natural language question $q$, so the answer $a$ is for future work.

This dataset is generated in a ``reverse question generation'' manner.
We first enumerate the pair $(n, f)$, then send the content of $f$ to GPT-4 to let it generate several pairs of $(q, a)$.
We prompt GPT-4 to ask questions on the functionality of the file by requiring it to analyze the file's role in the whole repository.

This dataset contains $1764$ training data ($292$ user queries), $505$ validation data ($86$ user queries), and $252$ test data ($44$ user queries).

Here is the prompt template for \autoexplore label generation:

\texttt{Below is a text file \{NAME\} from a repository. This repository is deployed as a backend service, providing users with certain services. Users want to use specific functionality or ask questions about the services, such as "tell me the business philosophy of this company" or "what is the high-level architecture of the proposed model". These inquiries are guaranteed to be answered by reading some text files.\\
    \\
    Your task is to first analyze its content. Then, come up with some user queries which involves this text file, along with the answers to them. Use the following format:\\
    \\
    \# ANALYSIS\\
    ...\\
    - QUERY1: ...\\
    - ANSWER1: ...\\
    - QUERY2: ...\\
    - ANSWER2: ...\\
    \\
    ----- Text -----\\
    \{CONTENT\}}

\subsection{Reflection Generation}

Here is the system message for \autoexplore reflection generation:

\texttt{You are a helpful assistant to explore a file system. Given a natural language task, you need to generate a sequence of system commands to identify the correct file. During interaction, you can only output a single choice number as response, which comes from a list of commands given to you. For example, the possible commands are: ["A. cat test.py", "c. cd progs", "9. cd .."]. Your answer should be "A", "c", or "9", not the entire command.\\ \\
    A special command `id X` is introduced to this task, which means to identify the file X as the final answer. Once you are sure X is the answer, use `id` to explicitly identify it, then the interaction terminates. Remember, simply `cat` a file does not identify it.}

Here is $p_{\textup{reflect}}$: \texttt{Now analyze the current situation and plan what to do next using 50 words. Don't give the choice yet. If you have identified the correct file in previous steps, you should exit at this step.}

Here is $p_{\textup{negative}}$: \texttt{The last command opens a wrong folder or file, which is a suboptimal move. Give reasons why the folder or file is incorrect and plan what to do next using 50 words. Don't give the choice yet.}

%% file: 9_appendix_exp.tex
\section{Other Benchmark Details}

\subsection{Dangerous Taxi}

OpenAI Gym uses MIT License.

Here is the system message for \taxi reflection generation:

\texttt{Given a problem state, the actions you have taken, and the observations you have.\\
\\
You need to give reflection on your actions, such as:\\
- What is the consequence of your previous action?\\
- How is your previous action? Good or bad? Why?\\
- What is the next action you want to take if possible? Why?\\
\\
I might give you some spoiler information and optimal action for cheating, but you should not mention that you have seen any spoilers, optimal actions, or any other information that you should not know.\\
Pretend you are smart and just know these information.\\
\\
Don't use any words related to "optimal" in your reflection.\\
\\
Keep your reflection concise within 100 words.\\
\\
For instance,\\
Because ..., so I ...\\
The task is to,... I\\
I found ... So...\\
etc.}

Here is $p_{\textup{negative}}$: \texttt{The previous actions might contain some mistakes.}
$p_{\textup{negative}}$ is directly appended to the observation prompt $p(s)$.

\subsection{ALFWorld}

ALFWorld and TextWorld use MIT License, Fast Downward uses GNU General Public License (GPL) v3.0.

\alfw shares the same system message and $p_{\textup{negative}}$ with \taxi.

\section{Experiment Details}

\begin{figure}[H]
    \centering
    \begin{subfigure}{.33\linewidth}
        \centering
        \fbox{\includegraphics[width=0.9\linewidth]{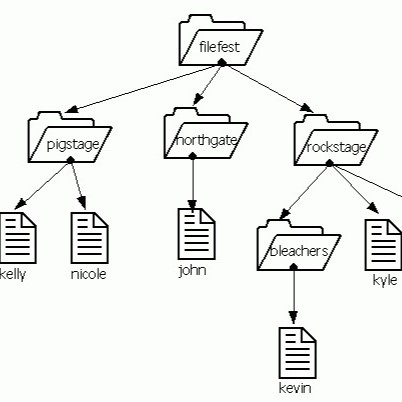}} 
        \caption{\tiny  \autoexplore}
    \end{subfigure}%
    \begin{subfigure}{.33\linewidth}
        \centering
        \fbox{\includegraphics[width=0.9\linewidth]{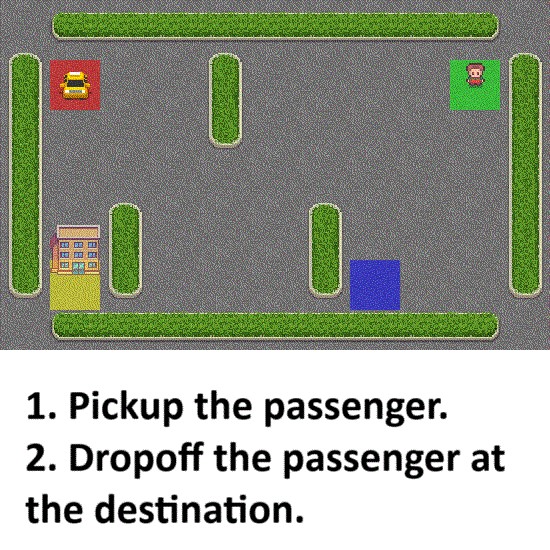}} 
        \caption{\tiny \taxi}
    \end{subfigure}%
    \begin{subfigure}{.33\linewidth}
        \centering
        \fbox{\includegraphics[width=0.9\linewidth]{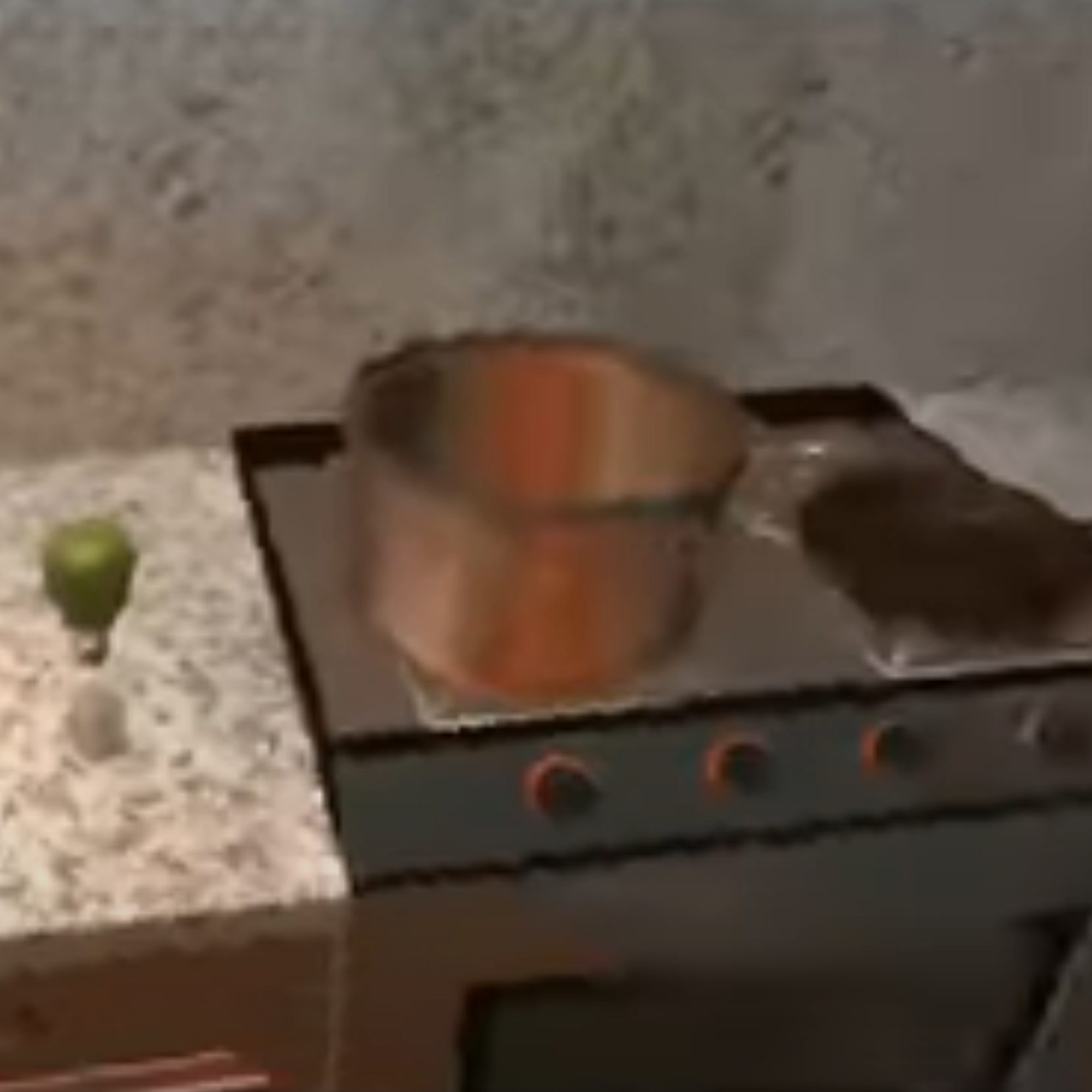}} 
        \caption{\tiny \alfw}
    \end{subfigure}
    \caption{Illustration of our environment}
    \label{fig:envs}
\end{figure}

We use NVIDIA RTX 4090, RTX A6000, and Tesla A40 for the training and evaluation of our proposed \rlft method. Python, PyTorch, HuggingFace PEFT, and AutoGen are used throughout the project.

All the experiments share the set of hyperparameters in \Cref{tab:hparam}.

\begin{table*}[t]
    \centering
    
    \begin{tabular}{l | l}
    \hline
     \textbf{Hyperparameter} & \textbf{Value}  \\
     \hline 
     \hline
        Train batch size & $1$ on 4090 \\
         & $2$ on A6000 and A40 \\
        Evaluate / Sample trajectory batch size & $4$ \\
        Gradient accumulation steps & $1$ \\
    \hline
        Learning rate & $2 \times 10^{-4}$ \\
        Gradient clipping norm & $0.3$ \\
        Weight decay & $1 \times 10^{-3}$ \\
     \hline
        Precision & bf$16$ \\
    \hline
        LoRA quantization & $8$bit \\
        LoRA $\alpha$ & $16$ \\
        LoRA rank & $64$ \\
     \hline
        Maximum token length & $1024$ \\
        Temperature & $1$ \\
        Top $p$ & $1$ \\
        Top $k$ & $99999$\\
     \hline
    \end{tabular}
    \caption{Hyperparameters of experiments}
    \label{tab:hparam}
\end{table*}

For \autoexplore, it takes around $30$ sec per iteration under the setting of horizon $H=15$.
For \taxi, it takes $30$ sec per iteration under the setting of horizon $H=15$ in pickup, and $60$ sec per iteration under the the setting of horizon $H=30$ in dropoff.
For \alfw, it takes $90$ sec per iteration under the setting of horizon $H=15$.

%% file: 9_appendix_qualitative_obs.tex
\definecolor{reflectionColor}{HTML}{FFA500} 
\definecolor{actionColor}{HTML}{0000FF} 

\newcommand{\displayLog}[4]{%
    \begin{mdframed}[linecolor=black,innertopmargin=5pt,innerbottommargin=5pt,leftmargin=0pt,rightmargin=0pt]
    \noindent\textbf{Observation:}\\\texttt{#1}\\
    \noindent\textbf{Reflection:}\\\textcolor{orange}{\texttt{#2}}\\
    \noindent\textbf{Action List:}\\\texttt{#3}\\
    \noindent\textbf{Action:}\\\textcolor{blue}{\texttt{#4}}
    \end{mdframed}
}

\section{Qualitative Observations}

\subsection{Illustration of Reflection-Aided Decision-Making} \label{sec:illustration_radm}

\Cref{fig:radm_1,fig:radm_2,fig:radm_3,fig:radm_4,fig:radm_5,fig:radm_6} show an entire trajectory of our reflection-aided decision-making method.
Irrelevant observations and actions are omitted by ``...''.

\begin{figure}[ht]
    \displayLog{You are in the middle of a room. Looking quickly around you, you see a cabinet 20, ..., a fridge 1, ...\\
    Your task is to: put a tomato in microwave.}{Begin by searching the cabinets or the fridge to find a tomato, then place it into the microwave. // Check cabinets: The most common place to find a tomato is in the cabinet or the fridge, but sometimes a cabinet isn't relevant.}{Action 0: go to cabinet 1\\
    Action 1: go to cabinet 10\\
    ...\\
    Action 27: go to fridge 1\\
    ...}{27}

    \caption{Step 1 of a reflection-aided decision-making example.}
    \label{fig:radm_1}
\end{figure}

\begin{figure}[ht]
    \displayLog{You are in the middle of a room. Looking quickly around you, you see a cabinet 20, ...\\
    Your task is to: put a tomato in microwave.\\
    Here is the previous path:\\
    --- Step: 0 ---\\
    Action: go to fridge 1\\
    Observation: You arrive at loc 24. The fridge 1 is closed.}{The initial action of going to the fridge aligns well with the task, as tomatoes are typically stored in a fridge. To progress, the next logical action would be to open the fridge to search for a tomato.}{Action 0: examine fridge 1\\
    Action 1: go to cabinet 1\\
    ...\\
    Action 41: open fridge 1}{41}

    \caption{Step 2 of a reflection-aided decision-making example.}
    \label{fig:radm_2}
\end{figure}

\begin{figure}[ht]
    \displayLog{You are in the middle of a room. Looking quickly around you, you see a cabinet 20, ...\\
    Your task is to: put a tomato in microwave.\\
    Here is the previous path:\\
    --- Step: 0 ---\\
    ...\\
    --- Step: 1 ---\\
    Action: open fridge 1\\
    Observation: You open the fridge 1. The fridge 1 is open. In it, you see a egg 1, a pot 2, a pot 1, a tomato 2, and a tomato 1.}{My initial actions were efficient, as I quickly located and opened the fridge to find the tomato necessary for my task. The next logical step is to take a tomato from the fridge and then proceed to put it in the microwave to complete the task.}{Action 0: close fridge 1\\
    ...\\
    Action 46: take tomato 2 from fridge 1}{46}

    \caption{Step 3 of a reflection-aided decision-making example.}
    \label{fig:radm_3}
\end{figure}

\begin{figure}[ht]
    \displayLog{You are in the middle of a room. Looking quickly around you, you see a cabinet 20, ..., a microwave 1, ...\\
    Your task is to: put a tomato in microwave.\\
    Here is the previous path:\\
    --- Step: 0 ---\\
    ...\\
    --- Step: 2 ---\\
    Action: take tomato 2 from fridge 1\\
    Observation: You pick up the tomato 2 from the fridge 1.}{The actions taken so far have effectively brought the tomato into my hands, which is in alignment with the first part of the given task. The next step is to navigate to the microwave and complete the operation to place the tomato inside it to fulfill the objective of placing a tomato in the microwave.}{Action 0: close fridge 1\\
    ...\\
    Action 32: go to microwave 1\\
    ...}{32}

    \caption{Step 4 of a reflection-aided decision-making example.}
    \label{fig:radm_4}
\end{figure}

\begin{figure}[ht]
    \displayLog{You are in the middle of a room. Looking quickly around you, you see a cabinet 20, ...\\
    Your task is to: put a tomato in microwave.\\
    Here is the previous path:\\
    --- Step: 0 ---\\
    ...\\
    --- Step: 3 ---\\
    Action: go to microwave 1\\
    Observation: You arrive at loc 22. The microwave 1 is closed.}{The actions taken thus far have successfully identified and retrieved the tomato from the fridge, progressing towards the task goal. The next step should be to open the microwave and put the tomato inside to complete the task effectively.}{Action 0: examine microwave 1\\
    ...\\
    Action 43: open microwave 1}{43}

    \caption{Step 5 of a reflection-aided decision-making example.}
    \label{fig:radm_5}
\end{figure}

\begin{figure}[ht]
    \displayLog{You are in the middle of a room. Looking quickly around you, you see a cabinet 20, ...\\
    Your task is to: put a tomato in microwave.\\
    Here is the previous path:\\
    --- Step: 0 ---\\
    ...\\
    --- Step: 4 ---\\
    Action: open microwave 1\\
    Observation: You open the microwave 1. The microwave 1 is open. In it, you see a mug 1.}{Successfully procuring the tomato from the fridge was a correct step towards completing the task; however, an oversight occurred in not removing the mug from the microwave before attempting to place the tomato inside. The next action should be to put the tomato in the microwave, which completes the task at hand.}{Action 0: close microwave 1\\
    ...\\
    Action 44: put tomato 2 in/on microwave 1}{44}

    \caption{Step 6 of a reflection-aided decision-making example.}
    \label{fig:radm_6}
\end{figure}

\subsection{Same Model for Reflection and Policy} \label{sec:same_model}

\Cref{fig:same_model_A,fig:same_model_B} are two log scripts in an experiment.
In this experiment, the reflection model $\wh{R}$ and policy model $\pi_{\theta_t}$ are the same.
Thus, the gradients computed on the action part will be influence the reflection part.
The reflect quality degraded drastically after 1194 iterations of updates.

\begin{figure}[ht]
    \displayLog{\# Task\\
    What is the reason for the planned domain name change for the website?\\
    \# Current working directory\\
    .\\
    \# Files under current working directory\\
    README.md\\
    \# Command history\\
    \\
    \# Execution result of your last command\\
    }{The README.md file typically contains information about the project, including reasons for the planned domain name change.}{t. exit\\
    103. cat README.md\\
    110. id README.md}{103}

    \caption{A log script from iteration \#10.}
    \label{fig:same_model_A}
\end{figure}

\begin{figure}[ht]
    \displayLog{\# Task\\
    What is the reason for the planned domain name change for the website?\\
    \# Current working directory\\
    .\\
    \# Files under current working directory\\
    README.md\\
    \# Command history\\
    \\
    \# Execution result of your last command\\
    }{cat README.md id README.md \# Explanation: cat README.md id README.md \# Explanation: cat README.MD  id README.md}{166. id README.md\\
    T. exit\\
    g. cat README.md\\}{g}

    \caption{A log script from iteration \#1204.}
    \label{fig:same_model_B}
\end{figure}